\documentclass[lettersize,journal]{IEEEtran}
\usepackage{amsmath,amsfonts}
\usepackage{array}
\usepackage[caption=false,font=normalsize,labelfont=sf,textfont=sf]{subfig}
\usepackage{textcomp}
\usepackage{stfloats}
\usepackage{url}
\usepackage{verbatim}
\usepackage{graphicx}
\usepackage{cite}
\usepackage{CJK}
\usepackage{makecell}
\usepackage{multirow}
\usepackage{enumitem}
\usepackage{algpseudocode}

\usepackage{amssymb}
\graphicspath{{authors/}} 
\usepackage{threeparttable}

\newcommand{\blue}[1]{\textcolor{blue}{#1}} 
\usepackage{pifont} 
\usepackage{booktabs}
\usepackage{pifont}
\usepackage[table]{xcolor}

\usepackage{xcolor}
\newcommand{\red}[1]{\textcolor{red}{#1}}
\usepackage{hyperref}
\usepackage{amssymb}

\usepackage{color}
\usepackage{xcolor}

\begin{document}

\title{An Underwater, Fault-Tolerant, Laser-Aided Robotic Multi-Modal Dense SLAM System for Continuous Underwater In-Situ Observation}

\author{Yaming Ou$^{1,2,3}$,~\IEEEmembership{Student Member,~IEEE,}
        Junfeng Fan$^1$,~\IEEEmembership{Senior Member,~IEEE,}
        Chao Zhou$^1$,~\IEEEmembership{Senior Member,~IEEE,}
        Pengju Zhang$^1$,
        Zongyuan Shen$^2$,
        Yichen Fu$^{1,3}$,
        Xiaoyan Liu$^{1,3}$,
        Zengguang Hou$^1$,~\IEEEmembership{Fellow,~IEEE}

\thanks{$^1$Institute of Automation, Chinese Academy of Sciences, Beijing, China.} 
\thanks{$^2$Robotics Institute, Carnegie Mellon University, Pittsburgh, USA.}
\thanks{$^3$School of Artificial Intelligence, University of Chinese Academy of Sciences, Beijing, China.}
}

\maketitle

\begin{abstract}
Existing underwater SLAM systems are difficult to work effectively in texture-sparse and geometrically degraded underwater environments, resulting in intermittent tracking and sparse mapping.
Therefore, we present Water-DSLAM, a novel laser-aided multi-sensor fusion system that can achieve uninterrupted, fault-tolerant dense SLAM capable of continuous in-situ observation in diverse complex underwater scenarios through three key innovations: 
Firstly, we develop Water-Scanner, a multi-sensor fusion robotic platform featuring a self-designed Underwater Binocular Structured Light (UBSL) module that enables high-precision 3D perception. 
Secondly, we propose a fault-tolerant triple-subsystem architecture combining:
1) DP-INS (DVL- and Pressure-aided Inertial Navigation System): fusing inertial measurement unit, doppler velocity log, and pressure sensor based Error-State Kalman Filter (ESKF) to provide high-frequency absolute odometry
2) Water-UBSL: a novel Iterated ESKF (IESKF)-based tight coupling between UBSL and DP-INS to mitigate UBSL's degeneration issues
3) Water-Stereo: a fusion of DP-INS and stereo camera for accurate initialization and tracking.
Thirdly, we introduce a multi-modal factor graph back-end that dynamically fuses heterogeneous sensor data.
The proposed multi-sensor factor graph maintenance strategy efficiently addresses issues caused by asynchronous sensor frequencies and partial data loss. 
Experimental results demonstrate Water-DSLAM achieves superior robustness (0.039~m trajectory RMSE and 100\% continuity ratio during partial sensor dropout) and dense mapping (6922.4 points/$m^3$ in 750~$m^3$ water volume, approximately 10 times denser than existing methods) in various challenging environments, including pools, dark underwater scenes, 16-meter-deep sinkholes, and field rivers. 
This work represents the first complete solution for long-duration underwater robotic dense in-situ observation.
Our project is available at \href{https://water-scanner.github.io/}{https://water-scanner.github.io/}.

\end{abstract}


\section{Introduction}
Underwater in-situ observation technology, as a critical tool in marine environmental research \cite{rife2006design,masmitja2020mobile}, overcomes the spatiotemporal limitations of traditional sampling methods by directly acquiring high-fidelity data in situ \cite{burns2024situ}. It demonstrates unique advantages in key applications such as underwater ecological monitoring and subsea pipeline inspection, where underwater robots \cite{kim2024development,baines2022multi} play a pivotal role as primary observation platforms.
However, complex underwater environments \cite{kim2013real} characterized by low light, sparse textures, structural degradation and strong disturbances pose significant challenges to their autonomy and endurance.
In particular, the robot’s simultaneous localization and mapping (SLAM) \cite{cadena2016past} capability is crucial to the success of observation missions.

Most existing underwater SLAM solutions rely on dead reckoning (DR) navigation systems \cite{qin2024fast}, but their cumulative errors cause significant drift over time, degrading localization accuracy. 
Vision-based underwater SLAM \cite{huang2020dual} performs well under ideal conditions but fails in dark or texture-sparse environments, often leading to localization breakdowns. 
Meanwhile, sonar-based bathymetric SLAM \cite{vial2024underwater} can construct depth maps in areas with distinct structural features, yet its applicability is limited to long-range coarse measurements due to the precision constraints and ambiguity of acoustic sensing. Consequently, such methods fall short of supporting fine-grained in-situ observations in complex scenarios.
These limitations collectively indicate that current SLAM technologies remain incapable of delivering stable and reliable solutions for continuous in-situ observation tasks in challenging underwater settings.
Therefore, developing a new SLAM approach capable of ensuring both large-scale localization stability and high-precision dense mapping remains an urgent challenge.

Recent advances in laser-aided active vision systems \cite{fan2023development} show great promise for underwater applications due to their superior penetration capability.
Laser-aided structured light systems have achieved remarkable results in 3D reconstruction, enabling high-precision measurements in low-light conditions \cite{gu2020three}. 
This breakthrough opens new possibilities for dense underwater observation. 
However, due to the influence of underwater structural degradation, structured light primarily relies on other localization methods for large-scale mapping. Its tight integration with SLAM systems has not yet been fully explored, and map consistency requires further improvement. Additionally, the use of low-frequency odometer information and common forms of static measurements makes it challenging to achieve dense mapping over large areas.

Furthermore, while single-sensor systems offer valuable environmental data, their performance often degrades in complex underwater settings due to limited accuracy and robustness \cite{rahman2022svin2}. 
To address this limitation, we propose a novel underwater multi-modal dense SLAM system that fuses data from an Inertial Measurement Unit (IMU), Doppler Velocity Log (DVL), Pressure Sensor (PS), stereo camera, and a custom-designed self-scanning Underwater Binocular Structured Light (UBSL). 
The proposed system adopts a robust multi-subsystem architecture capable of continuous in-situ observation under partial sensor failure, while dynamically correcting accumulated drift when reliable measurements are reintroduced. 
By effectively leveraging heterogeneous sensor data, this research advances large-scale, fault-tolerant dense SLAM capabilities in various challenging underwater scenarios.
This advance not only resolves critical gaps in underwater dense SLAM but also offers solid technical support for future underwater autonomous exploration and fine-gained in-situ observation tasks.

\subsection{Motivation}
Multi-sensor fusion SLAM demonstrates superior adaptability for underwater environments. 
However, existing solutions still struggle to achieve large-scale, high-precision, and dense in-situ observation in underwater challenging scenarios such as low-light conditions, sparse textures, and degraded structures.

\subsubsection{Localization}
Existing vision-based multi-sensor SLAM approaches often suffer from low localization accuracy or even complete failure in complex underwater environments, such as dark and sparse textures scenes. While sonar-based multi-sensor SLAM systems can work in dark scenes, they have inherent limitations such as low measurement resolution and elevation angle ambiguity, making them unsuitable for local fine observation and precise localization. More critically, when feature tracking fails, existing vision- and sonar-based SLAM frameworks require frequent reinitialization, severely degrading localization accuracy and disrupting continuous observation. Robust and accurate large-scale localization in challenging underwater scenarios remains an open problem.

\subsubsection{Mapping}
 Existing vision- and sonar-based SLAM systems primarily focus on localization and navigation, making it difficult to achieve high-precision underwater dense mapping. This limits their suitability for fine-scale underwater structure inspection and in-situ observation tasks. Although structured-light vision systems can generate high-accuracy point clouds, they often struggle to maintain mapping consistency in the structural degradation scene, which is common in underwater scenarios. The tight integration of these systems with other sensors, aimed at enhancing fault tolerance and ensuring more consistent mapping, remains an open area for further investigation. Moreover, limitations such as low-frequency odometry and fixed-line structured-light configurations further affect refined dense mapping.

\subsection{Contribution}
To enable underwater large-scale, dense in-situ observation, this work presents the following key contributions:
\begin{itemize}
    \item A novel underwater, fault-tolerant, laser-aided robotic multi-modal dense SLAM system, named Water-DSLAM, is firstly proposed for continuous underwater in-situ observation. The system is designed to fuse inertial, acoustic, pressure, and passive visual sensors, with a proprietary self-scanning UBSL module to provide a rare and effective solution for underwater dense SLAM. To the best of our knowledge, this is the first structured-light-based underwater multi-modal dense SLAM system capable of continuous, fine-grained in-situ observation.
    \item A fault-tolerant triple-subsystem front-end architecture is presented in Water-DSLAM, which innovatively mitigates the adverse effects of random external sensor faults caused by environmental changes on the SLAM process. The ESKF-based DP-INS subsystem is proposed to acquire high-frequency data, which is tightly coupled with the Water-Stereo subsystem to enhance performance in feature-sparse environments, and with the Water-UBSL subsystem to improve robustness in structurally degraded scenarios. By them, two rare constraints can be obtained.
    \item A multi-modal factor graph-based back-end is presented for Water-DSLAM, which dynamically fuses heterogeneous sensor data. The proposed multi-sensor factor graph maintenance strategy efficiently handles asynchronous sensor frequencies and partial data loss. The introduction of a fault detection module prevents the interference of anomalous data in mapping within challenging scenarios.
    \item To validate the effectiveness of Water-DSLAM, Water-Scanner is developed as a multi-sensor fusion robotic platform, equipped with IMU, DVL, PS, and Stereo sensors, along with a proprietary UBSL module to enable high-precision 3D perception. Experiments conducted in various challenging underwater environments, including pool tests, dark underwater settings, a 16-meter-deep sinkhole, and field river scenarios, demonstrate the superior underwater in-situ observation capabilities of Water-Scanner. Our proposed method outperforms state-of-the-art underwater multi-modal SLAM systems in terms of localization accuracy and dense mapping quality.
\end{itemize}

\section{Relative Works}
\subsection{Underwater Multi-modal SLAM}

In underwater environments, single-modality SLAM \cite{zhang2024fully} can provide basic pose estimation under ideal conditions, but its accuracy and robustness are limited. In complex scenarios, SLAM may fail entirely due to sensor issues like acoustic reverberation and poor visual clarity.
To overcome these limitations, multi-sensor fusion approaches have emerged as a research focus, with particular attention given to vision-inertial combinations. IMU can effectively compensate for the shortcomings of visual sensors in dynamic environments \cite{joshi2019experimental}. Classic visual-inertial SLAM framework VINS-Mono \cite{qin2018vins} has demonstrated excellent performance in terrestrial environments. Zhao et al. explored the challenges and solutions of applying VINS-Mono to underwater scenarios \cite{zhao2020detecting}, introducing ORB features and dark channel prior to significantly reduce total drift. Zhang et al. \cite{zhang2021open} proposed the FBUS-EKF framework, which utilizes Extended Kalman Filter to fuse stereo vision and IMU data, effectively mitigating the impact of underwater noise on localization.

To enhance robustness, researchers have integrated additional onboard information. Joshi et al. leveraged robotic kinematics for pose propagation to recover from VIO failures, improving localization continuity in dynamic underwater settings \cite{joshi2023sm, joshi2023hybrid}. Song et al. incorporated absolute positioning from a long baseline system to mitigate global drift in visual-inertial navigation \cite{song2023acoustic}. Vargas et al. fused DVL and IMU data to achieve acoustic odometry, improving SLAM stability in low-visibility, texture-sparse conditions \cite{vargas2021robust}. To address DR drift, Huang et al. introduced DVL velocity measurements into visual odometry, enhancing overall system accuracy \cite{huang2023tightly,huang2024visual}. 
Similarly, the recent work by Xu et al. proposed the AQUA-SLAM framework, which tightly fuses DVL, camera, and IMU sensors within a graph optimization framework, featuring an efficient online sensor calibration algorithm \cite{xu2025aqua}.
These efforts demonstrate that integrating auxiliary sensors significantly improves the robustness and precision of visual-inertial odometry systems.
However, maintaining reliable performance over extended periods in challenging underwater environments remains difficult.

In addition, the fusion of sonar and camera has become a key approach for underwater applications.
To compensate for poor visual quality, Xu et al. \cite{xu2020integrated} employed BlueRobotics Ping sonar to estimate seafloor distance, providing depth cues for partially tracked features and significantly improving localization reliability. Rahman et al. extended the OKVIS framework to integrate acoustic data \cite{rahman2018sonar}, and later developed SVIn2, a sonar-visual-inertial SLAM system utilizing scanning sonar to enhance both localization and mapping \cite{ rahman2019svin2,rahman2022svin2}. Yang et al. \cite{yang2022absolute} addressed monocular VO scale ambiguity using multi-view 2D sonar measurements, while Cardaillac et al. \cite{cardaillac2023camera} combined cameras with multi-beam forward-looking sonar for improved scale estimation and mapping. Silveira et al. \cite{silveira2015open} proposed a biologically inspired approach by fusing sonar, camera, and other modalities, extending the terrestrial RatSLAM to 3D underwater scenarios through marine mammal-inspired navigation strategies.

\subsection{Underwater Dense Mapping}
Underwater mapping has long been a highly challenging research topic, with current studies still primarily relying on acoustic and visual devices.

Sonar systems, with their long-range measurement capabilities, remain the most widely used for underwater mapping. Mallios et al. \cite{mallios2017underwater} equipped an AUV with dual high-performance mechanical imaging sonars to capture distance data from both horizontal and vertical cave surfaces for 3D reconstruction. To reduce sonar noise, DeBortoli et al. \cite{debortoli2018real} applied a CNN-based method for real-time frame selection and feature annotation, improving reconstruction quality. Given the ambiguity of sonar data, Westman et al. \cite{westman2019wide} introduced a generative model to estimate missing elevation angles.
Shen et al. \cite{shen2017autonomous,shen2021ct} achieved 3D terrain reconstruction in unknown underwater environments by fusing data from a multi-beam sonar, DVL, and IMU, and employing a novel 3D coverage path planning method called CT-CPP.
Despite the long-range advantage, sonar systems are prone to severe noise and ambiguity caused by reflection effects, which limits mapping accuracy and hinders fine-scale reconstruction.

In addition to sonar systems, passive visual mapping using cameras has also made significant progress in underwater research.
Ichimaru et al. \cite{ichimaru2019unified} achieved accurate reconstruction through refractive interfaces using a single camera. 
Ferrera et al. \cite{ferrera2021hyperspectral} proposed a novel 3D mapping approach that combines data from traditional RGB cameras, inertial navigation systems, and hyperspectral imaging cameras, generating high-precision 3D results with hyperspectral textures. 
Willners et al. \cite{willners2021robust} proposed a SLAM framework with autonomous relocation, enabling consistent global mapping despite feature loss.
Roberson et al. \cite{johnson2017high} combined AUVs and diver-operated stereo systems to produce the high resolution optical 3D map of an ancient underwater city.
While stereo vision enables 3D mapping, this method is heavily affected by environmental factors such as lighting and visibility, making it unsuitable for challenging underwater conditions.

Structured light systems have shown promise as active visual methods for underwater mapping, offering high precision thanks to the strong penetration of lasers \cite{fan2023development}. Roman et al. \cite{roman2010application} used a 532 nm laser and camera on an ROV to create high-resolution bathymetric maps. Bodenman et al. \cite{bodenmann2017generation} applied a fixed-line structured light system to study manganese nodules, while Bräuer-Burchardt et al. \cite{brauer2023underwater} combined VINS-Mono motion estimation with structured light for underwater archaeology.
More recently, Hitchcox et al. \cite{hitchcox2023improving} integrated the Insight Pro system with inertial navigation and loop closure to scan a shipwreck in 3D. However, such methods depend on external sensors for pose estimation, limiting mapping accuracy. Additionally, the commercial structured light systems cannot perform scene scanning, resulting in limited mapping density. 
To overcome these, Palomer et al. \cite{palomer2019inspection} developed a scanning structured light system on the Girona 500 AUV, improving mapping via point cloud alignment. Yet, the AUV had to remain nearly stationary to reduce distortion, making it inefficient and difficult to use in degraded or more complex areas.
As seen above, while structured light systems outperform sonar and stereo vision in dense mapping potential, current methods often rely on loosely coupled odometry, limiting their effectiveness in large-scale, consistent, and dense mapping.

In conclusion, although multi-sensor fusion has significantly improved the robustness of underwater SLAM, two critical limitations remain: (1) the difficulty in achieving dense and high-precision mapping makes it less suitable for observation tasks; and (2) limited adaptability hinders long-term, large-scale operation under challenging conditions such as feature-sparse and complete darkness underwater environments.
\begin{figure*}[!h]
\centering
\includegraphics[width=0.9\linewidth]{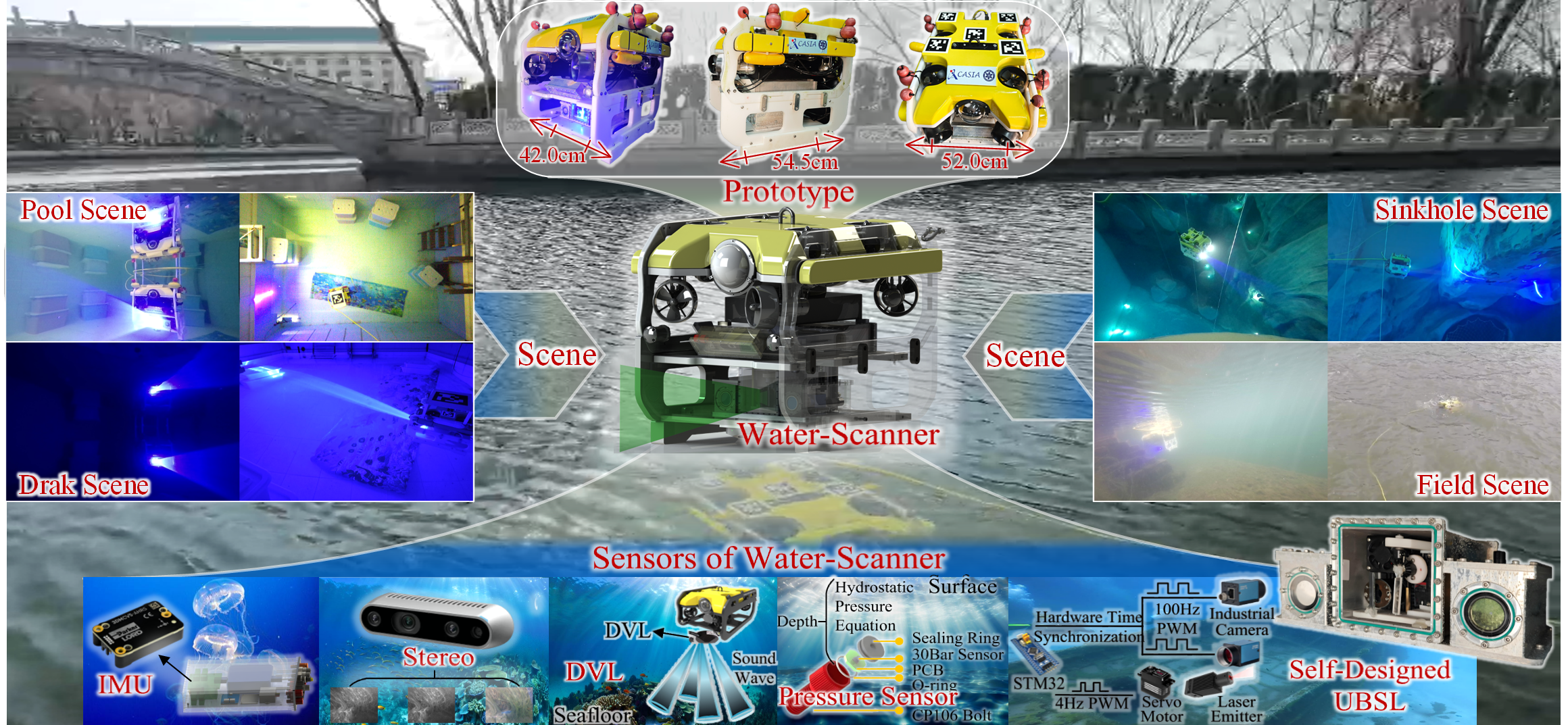}
\caption{{Mechatronic design, prototype, sensors and test scenarios of underwater multi-sensor robotic system, named Water-Scanner. Note: The picture in the background shows Water-Scanner's test site at Changhewan, a field river in Beijing, China.}}
\label{fig:system_hardware}
\end{figure*}
\section{Overview Of Water-Scanner System}
\subsection{Mechatronics Design of Water-Scanner}
To evaluate Water-DSLAM, we developed a multi-sensor underwater robotic system named Water-Scanner, designed for data collection in diverse underwater environments (Fig. \ref{fig:system_hardware}). The system comprises two main modules: the motion module and the sensor module.
The motion module, controlled by a Raspberry Pi 3B+, receives commands from a ground station via a neutrally buoyant cable (supporting Ethernet and 220 V power) and a micro switch. These commands are forwarded to a Pixhawk flight controller, which drives eight high-power thrusters for underwater maneuvering.
The sensor module is controlled by an NUC11TNH industrial computer. External sensors, including a PS, Stereo camera, underwater UBSL, and DVL, are waterproofed and connected to the NUC11TNH via cable glands. The internal IMU is directly linked to the NUC11TNH. PS data is collected by an STM32F407 module via I2C and transmitted to the NUC11TNH via USART. The Stereo camera and DVL directly transmit data via USB and USART, respectively.
Particularly, The custom-designed UBSL features two industrial cameras, a high-power laser, a servo motor, and two reflective mirrors. The cameras are synchronized using a 100~Hz PWM signal from the STM32F407. The servo oscillates one mirror ±45° at 2~Hz via variable PWM control. The laser emits a horizontal beam, which is reflected by the mirrors to perform horizontal scanning.
A point cloud from a single UBSL computation is called a \textbf{Scan} (70~Hz), while a point cloud from a full scan is called a \textbf{Sweep} (2~Hz).
Hardware specifications are listed in Table \ref{tab:hardware_para}. Additionally, the raspberry Pi 3B+ connects to the NUC11TNH via a network switch, establishing a shared LAN with the ground station for real-time sensor data monitoring.

\subsection{Sensors of Water-Scanner}
\begin{table}[!t]
\caption{System Sensor Parameters of Water-Scanner\label{tab:hardware_para}}
\centering
\begin{tabular}{ccc}
\Xhline{1.2pt}
\multicolumn{1}{c}{\textbf{Hardware}}          & \multicolumn{1}{c}{\textbf{Parameters}} & \multicolumn{1}{c}{\textbf{Value}}  \\ \Xhline{1.2pt}
\multirow{2}{*}{IMU} & Frequency             &   200 Hz             \\
                   & Communication Interface & USB \\\hline
\multirow{2}{*}{DVL} & Frequency             &   12 Hz             \\
                   & Communication Interface & USART \\\hline
{\multirow{2}{*}{PS}}&{Frequency}&{60 Hz}\\
& Resolving Power & 0.2 mbar \\\hline
\multirow{4}{*}{Stereo} & Resolution             &   640$\times$480 Pixels             \\
                   & Frames              &    30 Fps           \\ 
                   & FOV              &    $69.4^{\circ}\times42.5^{\circ}$         \\
                   & Communication Interface & USB 3.1 \\ \hline
\multirow{3}{*}{UBSL} & Frequency (Scan Pointcloud)             &   {70 Hz}             \\
& Frequency (Sweep Pointcloud)             &   {2 Hz}             \\
&No. of 3D points & 512 \\\hline

\multirow{3}{*}{UBSL: Laser} & Wavelength             &   450 nm             \\
                   &Power Consumption & 3 W \\
                   &Fan Angle & $35^\circ$\\ \hline
\multirow{4}{*}{UBSL: Camera} & Resolution             &   1280$\times$1024 Pixels             \\
                   & Frames              &    100 Fps           \\ 
                   & Focal Length              &    6 mm           \\
                   & Communication Interface & Ethernet \\
\Xhline{1.2pt}
\end{tabular}
\label{Tab_2_1}
\end{table}
To apply Water-DSLAM for continuous in-situ observation, Water-Scanner integrates some common underwater sensors as well as custom-developed sensors. These include the internal sensor IMU, and external sensors such as the acoustic sensor DVL, mechanical sensor PS, visual sensor Stereo camera, and a custom-designed laser-aided self-scanning UBSL. 
The characteristics of each sensor are as follows:
\begin{itemize}
\item \textbf{IMU} (Internal Sensor, Model: 3DM-CV5-25) — Provides high-frequency measurements of linear acceleration, angular velocity, and absolute orientation (magnetometer-assisted). Faults are not considered.
\item \textbf{DVL} (External Sensor, Model: WaterLinked A50) — Emits four acoustic beams towards the seabed and receives reflected signals to estimate the robot's velocity. In certain underwater environments, signal attenuation may lead to temporary sensor faults.  
\item \textbf{PS} (External Sensor, Model: AOHI B30) — Measures the pressure and temperature of the surrounding water to determine the absolute depth. Occasional anomalies in pressure readings may lead to temporary sensor faults.  
\item \textbf{Stereo} (External Sensor, Model: RealSense D435i) — Captures stereo images of the scene. In dark or texture-poor environments, Water-DSLAM tracking may fail, which is regarded as temporary sensor faults.  
\item \textbf{UBSL} (External Sensor, Custom-Developed) — Utilizes a refraction-based measurement model to obtain 3D information of the scene. It generates Scan point clouds per calculation and can produce sweep point clouds through continuous scanning. More details on the measurement principle can be found in recent work \cite{ou2023binocular}. The sweep point cloud generated from UBSL registration by Water-DSLAM may fail in regions with structural degradation, which is considered a temporary sensor fault.  
\end{itemize}
\begin{figure*}[!htbp]
\centering
\includegraphics[width=0.9\linewidth]{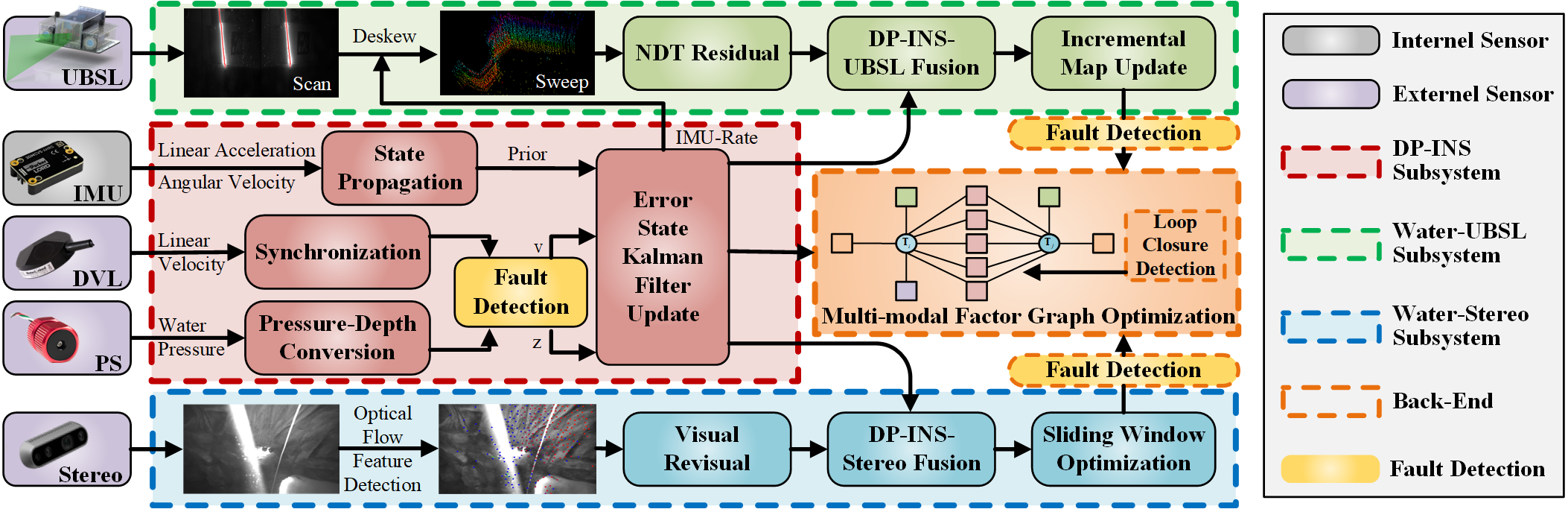}
\caption{{The proposed underwater, fault-tolerant, laser-aided robotic multi-modal dense SLAM framework for continuous underwater in-situ observation, named Water-DSLAM. It mainly consists of three parallel subsystems and a multi-modal factor graph back-end to achieve fault tolerance and continuous operation.}}
\label{fig:hardware}
\end{figure*}
\textbf{Notation:} This paper mainly focuses on the temporary faults of external sensors during the SLAM process. Since the data obtained by external sensors such as DVL, PS, Stereo, and UBSL are closely related to the underwater environment, these sensors often experience faults in complex underwater scenarios. Faults refer to abnormal or invalid data, distinguishing them from permanent damage.

\section{Overview of Water-DSLAM Framework}
Given the challenges of underwater environments, external sensors are prone to environmental disturbances and frequent failures. To address this, we propose Water-DSLAM, which is the first dense SLAM framework designed to handle harsh underwater conditions, enabling continuous in-situ observation even in darkness, low-texture areas, and degraded structures—despite frequent sensor disturbances.

Water-DSLAM employs a multi-subsystem architecture in the front-end, comprising the DP-INS, Water-UBSL, and Water-Stereo subsystems. 
Specifically, to meet the high-frequency distortion correction demands of structured light systems—unachievable by traditional low-rate DR systems ($<$10 Hz)—the DP-INS adopts an ESKF framework.
It propagates IMU states while correcting them with DVL and PS measurements, allowing temporary faults in both external sensors. 
This setup provides high-frequency absolute odometry, enhancing information utilization for other subsystems.
Generally, since the internal sensor IMU continuously provides valid data (although with increasing cumulative errors), occasional corrections from external sensors ensure robustness. This design underpins Water-DSLAM's uninterrupted, fault-tolerant operation. Unlike traditional SLAM systems reliant on external sensors like cameras or sonars, our architecture is better suited for real-world underwater deployment.

The Water-UBSL and Water-Stereo subsystems provide only relative constraints, avoiding typical underwater challenges such as structural degradation, feature sparsity and poor visibility. 
In Water-UBSL subsystem, UBSL data is tightly coupled with high-frequency data from the DP-INS subsystem, enabling distortion correction of high-rate Scan point clouds and enhancing robustness in point cloud registration, even in weakly structured environments. 
Similarly, the Water-Stereo subsystem tightly integrates DP-INS information to facilitate rapid initialization, avoiding poor initial estimates from erroneous PnP solutions that could lead to odometry failure. This integration also serves to constrain the internal windowed optimization within the Water-Stereo subsystem.

After passing through a fault detection module, the outputs of these three subsystems are fed into a multi-modal factor graph for back-end optimization. The designed multi-modal factor graph is capable of dynamically adjusting the connectivity graph in response to intermittent constraints from heterogeneous sensors. The proposed multi-sensor factor graph maintenance strategy effectively addresses challenges arising from asynchronous sensor frequencies and partial data faults.
To further enhance map consistency, the loop closure detection module integrates information from all three subsystems, enabling multi-modal loop detection and the computation of loop closure factors.

\section{Triple-subsystem Front-end of Water-DSLAM}

\subsection{DP-INS Subsystem in Water-DSLAM}

\begin{figure}[!t]
\centering
\includegraphics[width=0.9\linewidth]{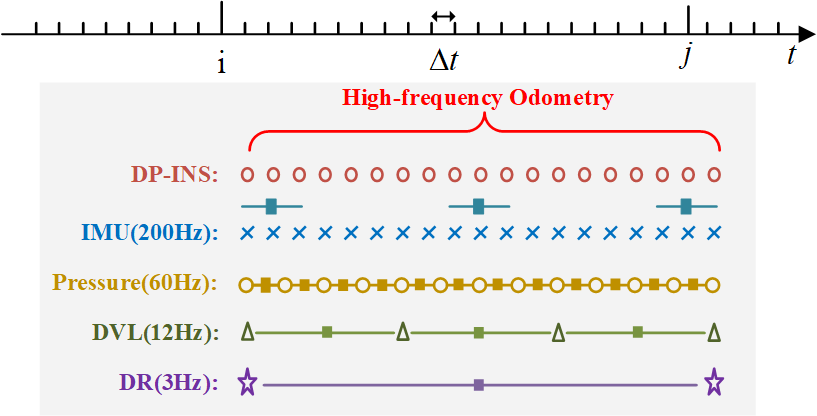}
\caption{{The frequency of sensors in DP-INS Subsystem.}}
\label{fig:DP-INS}
\end{figure}

\textbf{Objective: }The conventional underwater acoustic DVL WaterLinked A50 provides low-frequency DR data \cite{huang2024visual}, which suffers from two critical limitations: (1) its low update rate fails to match high-frequency laser scanning systems, leading to significant point cloud loss during mapping, and (2) its depth-axis velocity estimates exhibit substantial errors due to acoustic bottom-tracking. Therefore, we propose DP-INS based on the ESKF framework, which fuses IMU, DVL, and PS multi-modal information to realize IMU-rate robot motion estimation and greatly reduce the cumulative error of DVL DR.
Fig. \ref{fig:DP-INS} presents the frequency distribution of different sensor information.
Note: The following symbols without footnotes default to the IMU coordinate system.
$\boldsymbol{A}_\mathcal{B}^\mathcal{C}$  the conversion from the $\mathcal{B}$ coordinate system to the $\mathcal{C}$ coordinate system.

In DP-INS, let the truth state be $\boldsymbol{x}_\mathrm{t}=[\boldsymbol{p}_\mathrm{t},\boldsymbol{v}_\mathrm{t},\boldsymbol{R}_\mathrm{t},\boldsymbol{b}_\mathrm{gt},\boldsymbol{b}_\mathrm{at},\boldsymbol{g}_\mathrm{t}]^\top $, which is an 18-dimensional vector. 
It represents the robot position, velocity, and rotation matrices, IMU gyroscope bias, accelerometer bias, and gravity acceleration, in that order.
In continuous time, the relationship between the derivative of the state variable with respect to the IMU observations can be expressed as:
\begin{equation}
\begin{aligned}\dot{\boldsymbol{p}_\mathrm{t}}& =\boldsymbol{v}_{\mathrm{t}},  \\\dot{\boldsymbol{v}_{\mathrm{t}}}& =\boldsymbol{R}_\mathrm{t}(\tilde{\boldsymbol{a}}-\boldsymbol{b}_\mathrm{at}-\boldsymbol{\eta}_\mathrm{a})+\boldsymbol{g}_\mathrm{t},  \\\dot{\boldsymbol{R}_\mathrm{t}}& =\boldsymbol{R_\mathrm{t}}\left(\tilde{\boldsymbol{\omega}}-\boldsymbol{b_\mathrm{gt}}-\boldsymbol{\eta_\mathrm{g}}\right)^\wedge,  \\\dot{\boldsymbol{b}}_\mathrm{gt}& =\boldsymbol{\eta}_\mathrm{bg},\,\dot{\boldsymbol{b}}_\mathrm{at} =\boldsymbol{\eta}_\mathrm{ba},\,\dot{\boldsymbol{g}}_\mathrm{t}=\boldsymbol{0}. \end{aligned}
\label{eq:imu_observation}
\end{equation}
Where $\tilde{\boldsymbol{\omega}}$ and $\tilde{\boldsymbol{a}}$ are IMU linear acceleration and angular velocity readings.
Also, the nominal state in DP-INS is defined as $\boldsymbol{x}=[\boldsymbol{p},\boldsymbol{v},\boldsymbol{R},\boldsymbol{b}_\mathrm{g},\boldsymbol{b}_\mathrm{a},\boldsymbol{g}]^\top $ and the error state is defined as $\delta \boldsymbol{x}=[\delta \boldsymbol{p},\delta \boldsymbol{v},\delta \boldsymbol{R},\delta \boldsymbol{b}_\mathrm{g},\delta \boldsymbol{b}_\mathrm{a},\delta \boldsymbol{g}]^\top$, then
$
\boldsymbol{x}_\mathrm{t} = \boldsymbol{x} + \delta \boldsymbol{x}.
$

\subsubsection{Motion Prediction}
By taking the time derivative of above equation and discretizing the result, we obtain the discretized error-state equation:
\begin{equation}
    \begin{aligned}\delta \boldsymbol{p}\left(t+\Delta t\right)& =\delta \boldsymbol{p}+\delta\boldsymbol{v}\Delta t,  \\\delta\boldsymbol{v}(t+\Delta t)& =\delta\boldsymbol{v}+(-\boldsymbol{R}(\tilde{\boldsymbol{a}}-\boldsymbol{b}_\mathrm{a})^\wedge\delta\boldsymbol{\theta}-\boldsymbol{R}\delta\boldsymbol{b}_\mathrm{a}+\delta\boldsymbol{g})\Delta t-\boldsymbol{\eta}_\upsilon,  \\\delta\boldsymbol{\theta}(t+\Delta t)& =\mathrm{Exp}\left(-\left(\tilde{\boldsymbol{\omega}}-\boldsymbol{b}_{\mathrm{g}}\right)\Delta t\right)\delta\boldsymbol{\theta}-\delta\boldsymbol{b}_{\mathrm{g}}\Delta t-\boldsymbol{\eta}_{\theta},  \\\delta\boldsymbol{b}_\mathbf{g,a}(t+\Delta t)& =\delta\boldsymbol{b_\mathrm{g,a}}+\boldsymbol{\eta_\mathrm{g,a}},
    \delta \boldsymbol{g}\left(t+\Delta t\right)=\delta \boldsymbol{g}. \end{aligned}
\end{equation}
Then the motion prediction equation for DP-INS is obtained by linearizing the above discretized error state equation:
\begin{equation}
    \delta\boldsymbol{x}(t+\Delta t)=\underbrace{\boldsymbol{f}(\delta\boldsymbol{x}(t))}_{=\boldsymbol{0}}+\boldsymbol{F}\delta\boldsymbol{x}+\boldsymbol{w},\,\boldsymbol{\omega} \sim\mathcal{N}(0,\boldsymbol{Q}).
\end{equation}
Where,

\setlength{\arraycolsep}{2pt}  
\begin{equation}
    \begin{aligned}
    \boldsymbol{F} &= \begin{bmatrix}
    \boldsymbol{I} & \boldsymbol{I}\Delta t & \boldsymbol{0} & \boldsymbol{0} & \boldsymbol{0} & \boldsymbol{0} \\
    \boldsymbol{0} & \boldsymbol{I} & -\boldsymbol{R}(\tilde{\boldsymbol{a}}-\boldsymbol{b}_\mathbf{a})^{\wedge}\Delta t & \boldsymbol{0} & -\boldsymbol{R}\Delta t & \boldsymbol{I}\Delta t \\
    \boldsymbol{0} & \boldsymbol{0} & \operatorname{Exp}\left(-(\tilde{\boldsymbol{\omega}}-\boldsymbol{b}_\mathbf{g})\Delta t\right) & -\boldsymbol{I}\Delta t & \boldsymbol{0} & \boldsymbol{0} \\
    \boldsymbol{0} & \boldsymbol{0} & \boldsymbol{0} & \boldsymbol{I} & \boldsymbol{0} & \boldsymbol{0} \\
    \boldsymbol{0} & \boldsymbol{0} & \boldsymbol{0} & \boldsymbol{0} & \boldsymbol{I} & \boldsymbol{0} \\
    \boldsymbol{0} & \boldsymbol{0} & \boldsymbol{0} & \boldsymbol{0} & \boldsymbol{0} & \boldsymbol{I}
    \end{bmatrix}, \\
    \boldsymbol{Q} &= \mathrm{diag}(\boldsymbol{0}_3,\mathrm{Cov}(\boldsymbol{\eta}_v),\mathrm{Cov}(\boldsymbol{\eta}_\theta),\mathrm{Cov}(\boldsymbol{\eta}_g),\mathrm{Cov}(\boldsymbol{\eta}_a),\boldsymbol{0}_3).
    \end{aligned}
\end{equation}

\subsubsection{State Update}
In DP-INS, the observation model is expressed as:
\begin{equation}
\boldsymbol{z}=\boldsymbol{h}(\boldsymbol{x})+\boldsymbol{v},\,\boldsymbol{v}\sim \mathcal{N}(0,\boldsymbol{V}),
\end{equation}
where $\boldsymbol{z}$ is the measurement, $\boldsymbol{v}$ is the noise, and $\boldsymbol{V}$ its covariance. To update the nominal state $\boldsymbol{x}$ and error state $\delta\boldsymbol{x}$, we linearize the observation model and compute the Jacobian with respect to $\delta\boldsymbol{x}$:
\begin{equation}
\boldsymbol{H}=\left.\frac{\partial\boldsymbol{h}}{\partial\delta\boldsymbol{x}}\right|_{\boldsymbol{x}_\mathrm{pred}} = \frac{\partial \boldsymbol{h}}{\partial\boldsymbol{x}}\frac{\partial\boldsymbol{x}}{\partial\delta\boldsymbol{x}}.
\end{equation}
The second term is given by:
\begin{equation}
\frac{\partial\boldsymbol{x}}{\partial\delta\boldsymbol{x}}=\operatorname{diag}\left(\boldsymbol{I}_3,\boldsymbol{I}_3,\frac{\partial\operatorname{Log}(\boldsymbol{R}(\operatorname{Exp}(\delta\boldsymbol{\theta})))}{\partial\delta\boldsymbol{\theta}},\boldsymbol{I}_3,\boldsymbol{I}_3,\boldsymbol{I}_3\right).
\end{equation}
Then, the Kalman gain and state update are computed as:
\begin{equation}
\begin{aligned}&\boldsymbol{K}=\boldsymbol{P}_{\mathrm{pred}}\boldsymbol{H}^{\top}(\boldsymbol{H}\boldsymbol{P}_{\mathrm{pred}}\boldsymbol{H}^{\top}+\boldsymbol{V})^{-1},  \\&\delta \boldsymbol{x}=\boldsymbol{K}(\boldsymbol{z}-\boldsymbol{h}(\boldsymbol{x}_{\mathrm{pred}})),  \\&\boldsymbol{x}=\boldsymbol{x}_{\mathrm{pred}}+\delta\boldsymbol{x}, \,\,\boldsymbol{P}=\left(\boldsymbol{I}-\boldsymbol{K}\boldsymbol{H}\right)\boldsymbol{P}_{\mathrm{pred}}.\end{aligned}
\end{equation}

\subsubsection{DVL Velocity Observation}
The velocity observation model from the DVL is given by
$
\boldsymbol{v}_{\mathrm{dvl}} = \boldsymbol{R}^{\mathcal{D}}_{\mathcal{I}} \boldsymbol{v}.
$
Transforming the DVL measurement into the IMU coordinate system yields a direct observation of the velocity:
\begin{equation}
{\boldsymbol{R}^{\mathcal{D}}_{\mathcal{I}}}^\top \boldsymbol{v}_{\mathrm{dvl}} = \boldsymbol{v}.
\end{equation}
Clearly, its Jacobi matrix for the nominal state $\boldsymbol{x}$ is:
\begin{equation}
\frac{\partial \boldsymbol{h}(\boldsymbol{x})}{\partial \boldsymbol{x}} = 
\left[\boldsymbol{0}_{3\times3},\ \boldsymbol{I}_{3\times3},\ \boldsymbol{0}_{3\times12}\right].
\end{equation}
Thus, the DP-INS is updated with:
\begin{equation}
\boldsymbol{z} - \boldsymbol{h}(\boldsymbol{x}) = {\boldsymbol{R}^{\mathcal{D}}_{\mathcal{I}}}^\top \boldsymbol{v}_{\mathrm{dvl}} - \boldsymbol{v}.
\end{equation}

\subsubsection{PS Observation}
By converting the pressure value to depth value, the PS's observation is
$
    \boldsymbol{p}_\mathrm{ps} = [0,0,z_\mathrm{ps}].
$
The corresponding observation equation is:
$
    \boldsymbol{p}_\mathrm{ps}-\boldsymbol{p}_\mathrm{0} = \boldsymbol{R}_{\mathcal{I}}^{\mathcal{W}} \boldsymbol{p},
$
where $\boldsymbol{p}_\mathrm{0}$ is the absolute position obtained by the PS at the starting point, then it can be converted to:
\begin{equation}
{\boldsymbol{R}_{\mathcal{I}}^{\mathcal{W}}}^\top(\boldsymbol{p}_\mathrm{ps}-\boldsymbol{p}_\mathrm{0})=\boldsymbol{p}.
\end{equation}
Clearly, its Jacobi matrix for the nominal state $\boldsymbol{x}$ is:
\begin{equation}
    \frac{\partial\boldsymbol{h}(\boldsymbol(x))}{\partial\boldsymbol{x}}=[\boldsymbol{I}_{3\times3},\boldsymbol{0}_{3\times15}].
\end{equation}
So the DP-INS is updated with:
\begin{equation}
        \boldsymbol{z}-\boldsymbol{h}(\boldsymbol{x}) = {\boldsymbol{R}_{\mathcal{I}}^{\mathcal{W}}}^\top(\boldsymbol{p}_\mathrm{ps}-\boldsymbol{p}_0)-\boldsymbol{p}.
\end{equation}

\subsubsection{DR Observation}
Firstly, transfer the DR data to the IMU coordinate system:
\begin{equation}
    \begin{bmatrix}
\boldsymbol{R}^{\mathcal{I}}_{\mathcal{D}}  & \boldsymbol{p}^{\mathcal{I}}_{\mathcal{D}}\\
 \boldsymbol{0} &1
\end{bmatrix}
\begin{bmatrix}
\boldsymbol{R}_{\mathrm{DR} }  & \boldsymbol{p}_{\mathrm{DR} } \\
 \boldsymbol{0} &1
\end{bmatrix}
=
\begin{bmatrix}
\boldsymbol{R}^{\mathcal{I}}_{\mathcal{D}}\boldsymbol{R}_{\mathrm{DR} }  & \boldsymbol{R}^{\mathcal{I}}_{\mathcal{D}}\boldsymbol{t}_{\mathrm{DR} }+\boldsymbol{p}^{\mathcal{I}}_{\mathcal{D}} \\
 \boldsymbol{0} &1
\end{bmatrix}.
\end{equation}
Therefore, the DR-derived attitude satisfies $\boldsymbol{R}^{\mathcal{I}}_{\mathcal{D}}\boldsymbol{R}_{\mathrm{DR}} = \boldsymbol{R}\,\mathrm{Exp}(\delta\boldsymbol{\theta})$, where $\boldsymbol{R}$ is the nominal attitude and $\delta \boldsymbol{\theta}$ is the error. Since $\boldsymbol{R}$ is known, we define the observation as:
\begin{equation}
\boldsymbol{z_{\delta\theta}}=\boldsymbol{h}(\delta\boldsymbol{\theta})=\log_{\mathrm{SO(3)}}\left(\boldsymbol{R^\top \boldsymbol{R}^{\mathcal{I}}_{\mathcal{D}}R_\mathrm{DR}}\right)^\vee = \delta\boldsymbol{\theta}.
\end{equation}
where $\log_\mathrm{{SO(3)}}(\cdot)^\vee$ denotes the map from the three-dimensional rotation group SO(3) to the Lie-Algebra $\mathfrak{so}(3)$.
At this point, $\boldsymbol{z_{\delta\theta}}$ is a direct observation of $\delta\boldsymbol{\theta}$, and its Jacobian matrix with respect to $\delta\boldsymbol{\theta}$ is an identity matrix.

Similarly, for the translation observation equation
\begin{equation}
\boldsymbol{R}^{\mathcal{I}}_{\mathcal{D}}\boldsymbol{p}_{\mathrm{DR} }+\boldsymbol{p}^{\mathcal{I}}_{\mathcal{D}}=\boldsymbol{p}+\delta\boldsymbol{p},
\end{equation}
we can also change its form as follows:
\begin{equation}
    \boldsymbol{z_{\delta p}}=\boldsymbol{h}(\delta\boldsymbol{p})=\boldsymbol{R}^{\mathcal{I}}_{\mathcal{D}}\boldsymbol{p}_{\mathrm{DR} }+\boldsymbol{p}^{\mathcal{I}}_{\mathcal{D}}-\boldsymbol{p} = \delta\boldsymbol{p}.
\end{equation}
So the Jacobi matrix with respect to $\delta\boldsymbol{p}$ is also an identity matrix.
Note that the original update variable $\boldsymbol{z}-\boldsymbol{h}(\boldsymbol{x})$ in DP-INS should be transformed to:
\begin{equation}
    \boldsymbol{z}-\boldsymbol{h}(\boldsymbol{x})=[\boldsymbol{R}^{\mathcal{I}}_{\mathcal{D}}\boldsymbol{p}_{\mathrm{DR} }+\boldsymbol{p}^{\mathcal{I}}_{\mathcal{D}}-\boldsymbol{p},\mathrm{Log}(\boldsymbol{R}^\top\boldsymbol{R}^{\mathcal{I}}_{\mathcal{D}}\boldsymbol{R}_\mathrm{DR})]^\top.
\end{equation}
In summary, the Jacobi matrix of DR observations is:
\begin{equation}
    \frac{\partial\boldsymbol{h}(\boldsymbol(\delta x))}{\partial\boldsymbol{\delta x}}=[\boldsymbol{I}_{3\times3},\boldsymbol{0}_{3\times3},\boldsymbol{I}_{3\times3},\boldsymbol{0}_{3\times9}].
\end{equation}

\subsubsection{Fault Detection}
Due to the challenging underwater environment, measurements from the DVL and PS may contain outliers. To mitigate this, a statistical fault detection method is adopted. Let the current measurement be $\mathbf{M}$, and the historical data within a sliding window of size $n$ be $D = \{m_1, m_2, \ldots, m_n\}$. Denote the mean and standard deviation of $D$ as $\mathbf{e}$ and $\mathbf{s}$, respectively. A measurement is classified as an outlier and rejected if
$
    \|\mathbf{M} - \mathbf{e}\| > \xi \|\mathbf{s}\|.
$
The threshold $\xi$ is commonly set to 3 in practice, but may be adjusted according to empirical requirements.

\subsection{Tightly-coupled Water-UBSL Subsystem}
\textbf{Objective: }Structural degradation and intense robot shaking caused by water disturbances in challenging underwater environments often lead the loosely coupled UBSL method in \cite{ou2023water} to fail in point cloud matching during large-scale movement. To mitigate these issues, we propose the Water-UBSL subsystem, a tightly-coupled system based on the IESKF framework that integrates UBSL and DP-INS information to enhance robustness. It aims to estimate the robot's relative pose transformation between two Sweep point cloud obtained from UBSL, thereby providing additional constraints for back-end optimization and improving map consistency.

\begin{figure}[!t]
\centering
\includegraphics[width=1\linewidth]{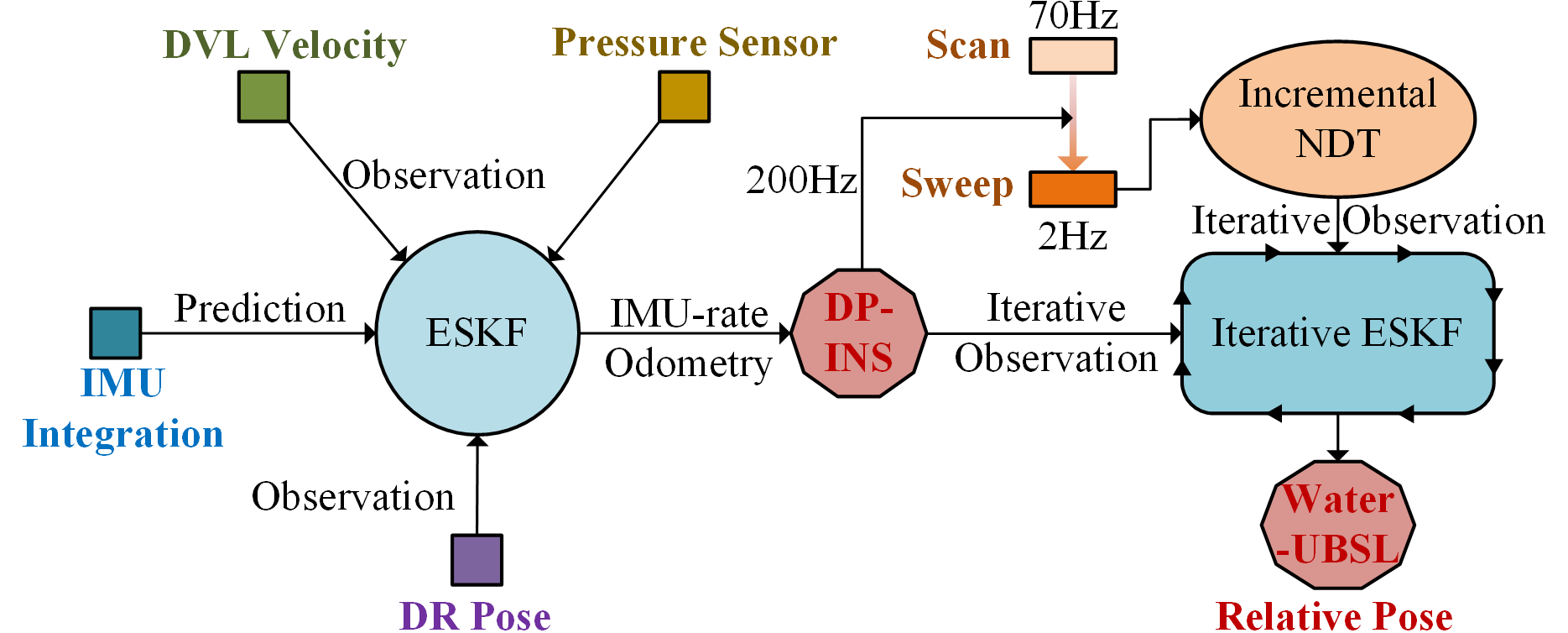}
\caption{{The designed schematic of Water-UBSL Subsystem.}}
\label{fig:water-ubsl}
\end{figure}

The custom-developed UBSL acquires Scan point clouds at approximately 70 Hz. Constructing a Sweep point cloud for odometry estimation requires associating each Scan with the robot’s corresponding state. Assuming the robot remains stationary during scanning, as in \cite{palomer2019inspection}, introduces significant errors. With high-frequency odometry from the DP-INS subsystem, multiple Scans can be accurately integrated into a Sweep point cloud, as shown in Fig. \ref{fig:water-ubsl}.

To obtain UBSL constraint factors, we perform Sweep point cloud registration using the Normal Distributions Transform (NDT), which enhances robustness by modeling the spatial distribution of points statistically.
Each Sweep is voxelized, and a Gaussian distribution characterized by mean $\boldsymbol{\mu}_k$ and covariance $\boldsymbol{\Sigma}_k$ is computed for each voxel.
Given a point $\boldsymbol{q}_i$ from the source cloud, after applying the current estimated transformation $(\boldsymbol{R}, \boldsymbol{t})$, it falls into a voxel with parameters $\boldsymbol{\mu}_i$ and $\boldsymbol{\Sigma}_i$. The residual for this voxel is defined as:
\begin{equation}
        \boldsymbol{e}_i = \boldsymbol{R}\boldsymbol{q}_i + \boldsymbol{t} - \boldsymbol{\mu}_i.
\end{equation}
Thus, the optimization of $\boldsymbol{R},\boldsymbol{t}$ in each iteration is formulated as a weighted least squares problem:
\begin{equation}
        (\boldsymbol{R,t})^* = \arg \min_{\boldsymbol{R,t}} \sum_{i}(\boldsymbol{e}^T_i\boldsymbol{\sum}^{-1}_{i}\boldsymbol{e}_i).
\end{equation}
This can be solved using the Gauss-Newton method:
\begin{equation}
    \sum_{i}(\boldsymbol{J}^{T}_{i}{\sum}^{-1}_{i}\boldsymbol{J}_{i})\Delta \boldsymbol{x} = -\sum_{i}\boldsymbol{J}^{T}_{i}{\sum}^{-1}_{i}\boldsymbol{e}_{i},
\end{equation}
where $\Delta \boldsymbol{x}$ is the increment for each iteration, and $\boldsymbol{J}_{i}$ is the Jacobian matrix, given by:
\begin{equation}
\frac{\partial \boldsymbol{e}_i}{\partial \boldsymbol{R}} = -\boldsymbol{R}\boldsymbol{q}_{i}^\wedge,\, \frac{\partial \boldsymbol{e}_i}{\partial \boldsymbol{t}} = \boldsymbol{I}.
\label{eq:ji1}
\end{equation}

To improve Sweep point cloud matching efficiency, the Water-UBSL subsystem uses an incremental NDT approach for local map maintenance. Rather than rebuilding the local map for each new Sweep, the system simply adds the current Sweep to existing NDT voxels while removing old ones, keeping the total voxel count constant.
The voxel's Gaussian distribution updates as follows:
\begin{equation}
\begin{aligned}
&\boldsymbol{\Sigma}    = \frac{m(\boldsymbol{\Sigma_H}\!\! +\!\! (\boldsymbol{\mu_H}\!\! -\!\! \boldsymbol{\mu})(\boldsymbol{\mu_H}\!\! -\!\! \boldsymbol{\mu})^{\top})\!\! +\!\! n(\boldsymbol{\Sigma_A}\!\! + \!\!(\boldsymbol{\mu_A} \!\!-\!\! \boldsymbol{\mu})(\boldsymbol{\mu_A}\!\! -\!\! \boldsymbol{\mu})^{\top})}{m + n},\\
&\mu={(m \mu_{\mathrm{H}}+n \mu_{\mathrm{A}})}/{(m+n)},
\end{aligned}
\end{equation}
where $\boldsymbol{\Sigma_H}$ and $\mu_{\mathrm{H}}$ represent the mean and covariance of the historical points, while $\boldsymbol{\Sigma_A}$ and $\mu_{\mathrm{A}}$ correspond to the newly added points. $m$ and $n$ are the numbers of historical points and newly added points, respectively.

Due to the sparse underwater structural features, direct Sweep point cloud matching often fails. Therefore, we tightly integrate DP-INS data with UBSL within the IESKF framework, resulting in a more robust Water-UBSL subsystem.

According to the derivation in \cite{xu2021fast}, let $\boldsymbol{P}_{k}=\boldsymbol{J}_k\boldsymbol{P}_{i}\boldsymbol{J}^{T}_{k}$, the iterative update process for the $(i+1)$-th observation is:
\begin{equation}
    \begin{aligned}
&\boldsymbol{K}_{k}=\boldsymbol{P}_{k}\boldsymbol{H}_{k}^{\top}(\boldsymbol{H}_{k}\boldsymbol{P}_{k}\boldsymbol{H}_{k}^{\top}+\boldsymbol{V})^{-1},\\
&\delta \boldsymbol{x}_{k}=\boldsymbol{K}_{k}(\boldsymbol{z}-\boldsymbol{h}(\boldsymbol{x}_{k})).
    \end{aligned}
\label{ieskf}
\end{equation}
Here, $\boldsymbol{P}_{i}$ represents the covariance matrix from the previous observation, and $\boldsymbol{J}_k$ is the Jacobian matrix of the tangent space transformation for the $k$-th iteration.
According to the above, the nominal state $\delta \boldsymbol{x}_{k}$ and $\boldsymbol{J}_{k}$ are iteratively updated until convergence, after which the covariance $\boldsymbol{P}_{i+1}$ is updated as follows:
\begin{equation}
    \boldsymbol{P}_{i+1}=(\boldsymbol{I}-\boldsymbol{K}_{k}\boldsymbol{H}_{k})\boldsymbol{J}_{k}\boldsymbol{P}_{i}\boldsymbol{J}_{k}^{\top}.
\label{ieskf_p}
\end{equation}
By substituting this result into Eq. \ref{ieskf} and Eq. \ref{ieskf_p}, the iterative update of the IESKF at each step is obtained as:
\begin{equation}
    \begin{aligned}
    \delta \boldsymbol{x}_{k}&=(\boldsymbol{P}_k^{-1}+\boldsymbol{H}_k^\top \boldsymbol{V}^{-1}\boldsymbol{H}_k)^{-1}\boldsymbol{H}_k^\top \boldsymbol{V}^{-1}(\boldsymbol{z}-\boldsymbol{h}(\boldsymbol{x}_{k})),\\
    \boldsymbol{P}_{i+1}&=(\boldsymbol{I}\!\!-\!\!(\boldsymbol{P}_{k}^{-1}\!\!+\!\!\boldsymbol{H}_{k}^{\top}\boldsymbol{V}^{-1}\boldsymbol{H}_{k})^{-1}\boldsymbol{H}_{k}^{\top}\boldsymbol{V}^{-1}\boldsymbol{H}_{k})J_{k}\boldsymbol{P}_{i}J_{k}^{\top}.
    \end{aligned}
\label{eq: update}
\end{equation}
Moreover, we have $\boldsymbol{P}_{k} = \boldsymbol{J}_k \boldsymbol{P}_{i} \boldsymbol{J}^{T}_{k}$. 
Let 
\begin{equation}
\boldsymbol{Q}_{k} = \boldsymbol{H}_k^\top \boldsymbol{V}^{-1} \boldsymbol{H}_k \,,\, \boldsymbol{Z}_{k} = \boldsymbol{H}_k^\top \boldsymbol{V}^{-1} (\boldsymbol{z} - \boldsymbol{h}(\boldsymbol{x}_{k})), 
\label{eq: qkzk}
\end{equation}
then Eq. \ref{eq: update} becomes:
\begin{equation}
    \begin{aligned}
    \delta \boldsymbol{x}_{k}&=((\boldsymbol{J}_k\boldsymbol{P}_{i}\boldsymbol{J}^{T}_{k})^{-1}+\boldsymbol{Q}_k)^{-1}\boldsymbol{Z}_k,\\
    \boldsymbol{P}_{i+1}&=(\boldsymbol{I}-((\boldsymbol{J}_k\boldsymbol{P}_{i}\boldsymbol{J}^{T}_{k})^{-1}\!\!+\!\!\boldsymbol{Q}_k)^{-1}\boldsymbol{Q}_k)J_{k}\boldsymbol{P}_{i}J_{k}^{\top},
    \end{aligned}
\label{eq: update2}
\end{equation}
where $\boldsymbol{J}_k$ can be calculated from the state variable $\boldsymbol{x}_k$ at each iteration. For different observations, the main task is to derive $\boldsymbol{Q}_k$ and $\boldsymbol{Z}_k$.

\subsubsection{DP-INS Observation}
Similar to the previous DR Observation update process, the Jacobian matrix for the DP-INS Observation is given by:
\begin{equation}
    \boldsymbol{H}_k = \frac{\partial\boldsymbol{h}(\boldsymbol(\delta x))}{\partial\boldsymbol{\delta x}}=[\boldsymbol{I}_{3\times3},\boldsymbol{0}_{3\times3},\boldsymbol{I}_{3\times3},\boldsymbol{0}_{3\times9}].
\end{equation}
The covariance matrix $\boldsymbol{V}$ corresponds to the noise matrix of the DP-INS system. Additionally,
\begin{equation}
    \boldsymbol{z}-\boldsymbol{h}(\boldsymbol{x}_k)=[\boldsymbol{p}_{\mathrm{DP-INS} }-\boldsymbol{p}_k,\log_{\mathrm{SO(3)}}(\boldsymbol{R}_k^\top\boldsymbol{R}_\mathrm{DP-INS})^\vee]^\top.
\end{equation}
According to Eq. \ref{eq: qkzk}, the values of $\boldsymbol{Q}_k$ and $\boldsymbol{Z}_k$ for each iteration can be computed.

\subsubsection{Sweep NDT Observation}
According to the definition of the state variables in the IESKF framework and combining with Eq. \ref{eq:ji1}, for NDT matching of a Sweep containing $N$ points, the Jacobian matrix $\boldsymbol{J}_{j}$ corresponding to the residual $e_j$ of the $j$-th point is:
\begin{equation}
\begin{aligned}
    \boldsymbol{J}_j & = [\frac{\partial{\boldsymbol{e}_j}}{\partial{\boldsymbol{t}}},\boldsymbol{0}_3,\frac{\partial{\boldsymbol{e}_j}}{\partial{\boldsymbol{R}}},\boldsymbol{0}_{3\times3}] = [\boldsymbol{I}_3,\boldsymbol{0}_3,-\boldsymbol{R}\boldsymbol{q}_{i}^\wedge,\boldsymbol{0}_{3\times3}].
\end{aligned}
\end{equation}
The observation model Jacobian matrix and noise matrix are:
\begin{equation}
\boldsymbol{H}_k = [\boldsymbol{J}_1^\top \ \cdots\ \boldsymbol{J}_j^\top \ \cdots\ \boldsymbol{J}_N^\top]^\top, 
\boldsymbol{V} = \text{diag}(\boldsymbol{\Sigma}_1, \cdots, \boldsymbol{\Sigma}_N).
\end{equation}
Since $\boldsymbol{V}$ is a block-diagonal matrix, substituting into the Eq. \ref{eq: qkzk} yields:
\begin{equation}
   \boldsymbol{Q}_k = \sum_{j=1}^{N}\boldsymbol{J}_{j}^{\top}\boldsymbol{\Sigma}_{j}^{-1}\boldsymbol{J}_{j},
   \boldsymbol{Z}_k = \sum_{j=1}^{N}\boldsymbol{J}_{j}^{\top}\boldsymbol{\Sigma}_{j}^{-1}\boldsymbol{e}_{j}. 
\end{equation}

\subsection{DP-INS Aided Water-Stereo Subsystem}
\begin{figure}[!t]
\centering
\includegraphics[width=1\linewidth]{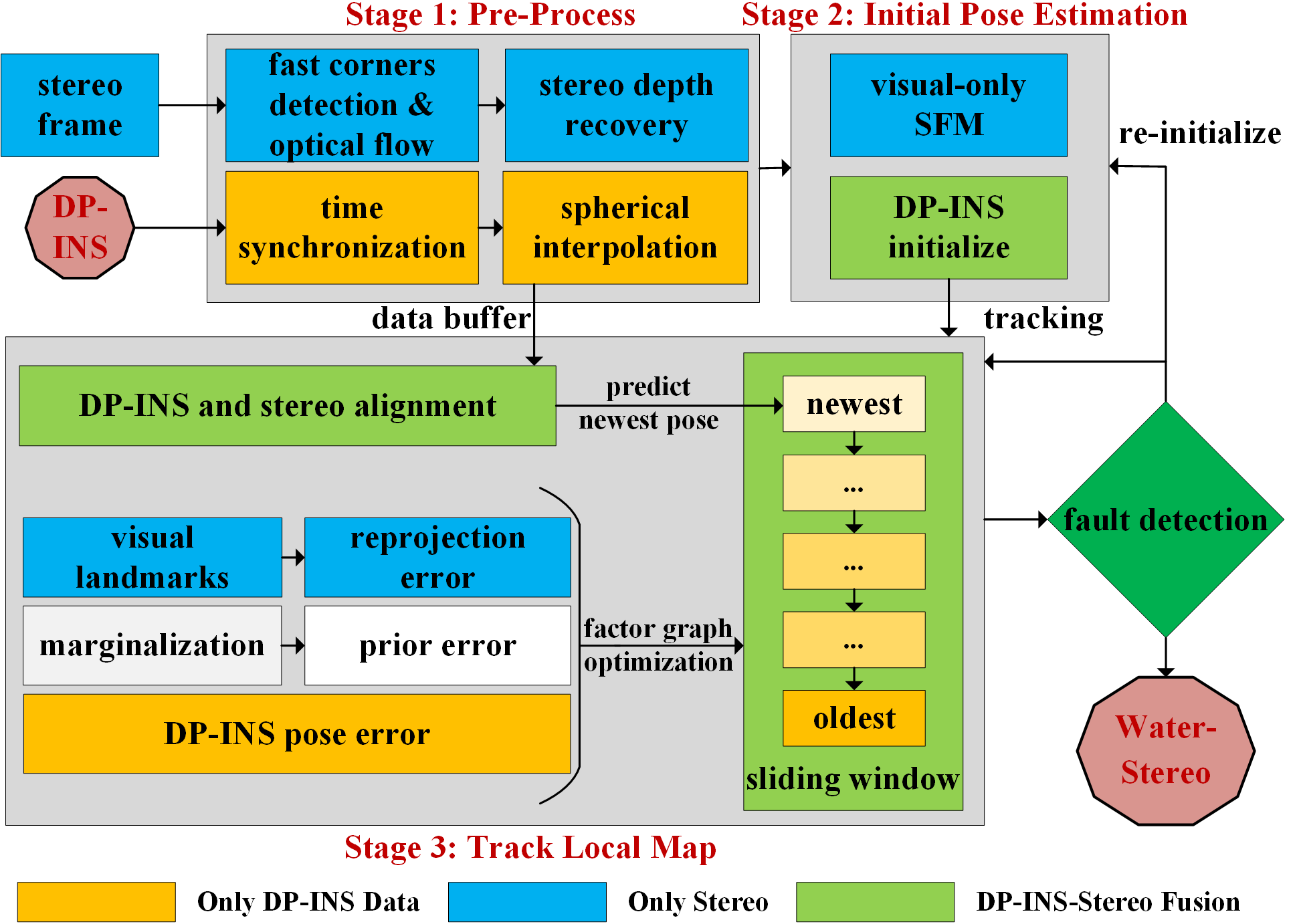}
\caption{{The designed schematic of Water-Stereo Subsystem.}}
\label{fig:water-stereo}
\end{figure}
\textbf{Objective: }Challenging underwater scenes often suffer from lighting changes, particle interference, and sparse textures, leading to unstable visual feature tracking and degraded SLAM performance. To improve robustness, we propose the Water-Stereo subsystem, which leverages motion priors from the DP-INS into VINS-Fusion \cite{qin2019general} for fast initialization and constrained visual optimization, as shown in Fig. \ref{fig:water-stereo}. This tightly coupled design produces reliable relative constraints that enhance map consistency in the Water-DSLAM back-end.

In the Water-Stereo, a DP-INS-Visual tightly-coupled factor graph optimization framework is proposed. 
The state variables in per frame are defined as follows:
\begin{equation}
\boldsymbol{x}=\left[\begin{array}{llllllll}
\boldsymbol{R}^T & \boldsymbol{p}^T  & \boldsymbol{\lambda}_1 &,...,&\boldsymbol{\lambda}_m
\end{array}\right],
\end{equation}
where $\boldsymbol{R}$, $\boldsymbol{p}$ denote the rotation and position of the state in the world frame. 
The $\boldsymbol{\lambda}_m$ denotes the inverse depth corresponding to the $m$th passive visual feature on the current left camera frame.
Then the sliding window optimization using multi-frame state variables together is presented below:
\begin{equation}
    {\mathcal{X}} = \left[\begin{array}{lllll}
\boldsymbol{x}_1 & \boldsymbol{x}_2 & \boldsymbol{x}_3 &,...,& \boldsymbol{x}_n
\end{array}\right]{,}
\end{equation}
where $n$ indicates the window size.
Various constraint factors are used to optimize these state variables.

\subsubsection{Stereo Factor}
For the left image frame \( c_i \) in {Water-Stereo}, a subset of matched features with \( c_j \) obtains high-precision map points \( \hat{\mathcal{D}}^{c_i}_l \ (l = 1, \dots, m) \) through stereo or active vision (UBSL), as in recent work~\cite{ou2024hybrid}. The remaining features \( [\hat{u}_l^{c_i}, \hat{v}_l^{c_i}] \ (l = 1, \dots, n) \), triangulated passively with inverse depth \( \lambda^{c_i}_l \), are represented as map points \( \hat{\mathcal{P}}^{c_i}_l([\hat{u}_l^{c_i}, \hat{v}_l^{c_i}, \lambda^{c_i}_l]) \). The stereo factor thus includes two components, with the first being the reprojection constraint
\begin{equation}
\boldsymbol{r}_{\mathcal{V}_1}(\mathcal{L}_i, \mathcal{R}_i, \mathcal{X}) = \sum_{l=1}^{n} ( f(\hat{\mathcal{P}}_l^{c_j^l}) - \frac{f(\hat{\mathcal{P}}_l^{c_i^l})}{\|f(\hat{\mathcal{P}}_l^{c_i^l})\|} + f(\hat{\mathcal{P}}_l^{c_j^r}) - \frac{f(\hat{\mathcal{P}}_l^{c_i^r})}{\|f(\hat{\mathcal{P}}_l^{c_i^r})\|} ),
\end{equation}
where
\begin{equation}
\left\{\begin{array}{l}
f(\hat{\mathcal{P}}_l^{c^*_j})=\pi_e{ }^{-1}\cdot[\begin{array}{c}
\hat{u}_l^{c^*_j},
\hat{v}_l^{c^*_j}
\end{array}]^ \top\\
\begin{aligned}
f(\hat{\mathcal{P}}_l^{c^*_i})=\boldsymbol{R}_b^c(\boldsymbol { R } _ { w } ^ { b _ { j } } (\boldsymbol { R } _ { b _ { i } } ^ { w } (\boldsymbol{R}_c^b ({\lambda^{c^*_i}_{l}})^{-1} \pi_c^{-1}([\begin{array}{c}
\hat{u}_l^{c^*_i},
\hat{v}_l^{c^*_i}
\end{array}]^\top)\\
+\boldsymbol{p}_c^b)+\boldsymbol{p}_{b_i}^w-\boldsymbol{p}_{b_j}^w)-\boldsymbol{p}_c^b).
\end{aligned}\\
\end{array}\right.
\end{equation}
Among them, \( \pi_e^{-1}(\cdot) \) represents the back-projection function, and \( c^* \) refers to the left and right cameras \( c^l, c^j \), respectively. Additionally, the second part of the stereo vision factor corresponds to the matched point depth factor for the left camera
\begin{equation}
\boldsymbol{r}_{\mathcal{V}_2}(\left(\mathcal{L}_{i},\hat{\mathcal{D}}^{c^r_i},\mathcal{X}\right))=\sum_{l=1}^{m} (g(\hat{\mathcal{D}}^{c^r_i}_{l}) - \lambda_{l}^{c^l_i}),
\end{equation}
where $g(\cdot)$ denotes the calculation of the inverse depth.
\begin{figure*}[!htbp]
\centering
\includegraphics[width=0.95\linewidth]{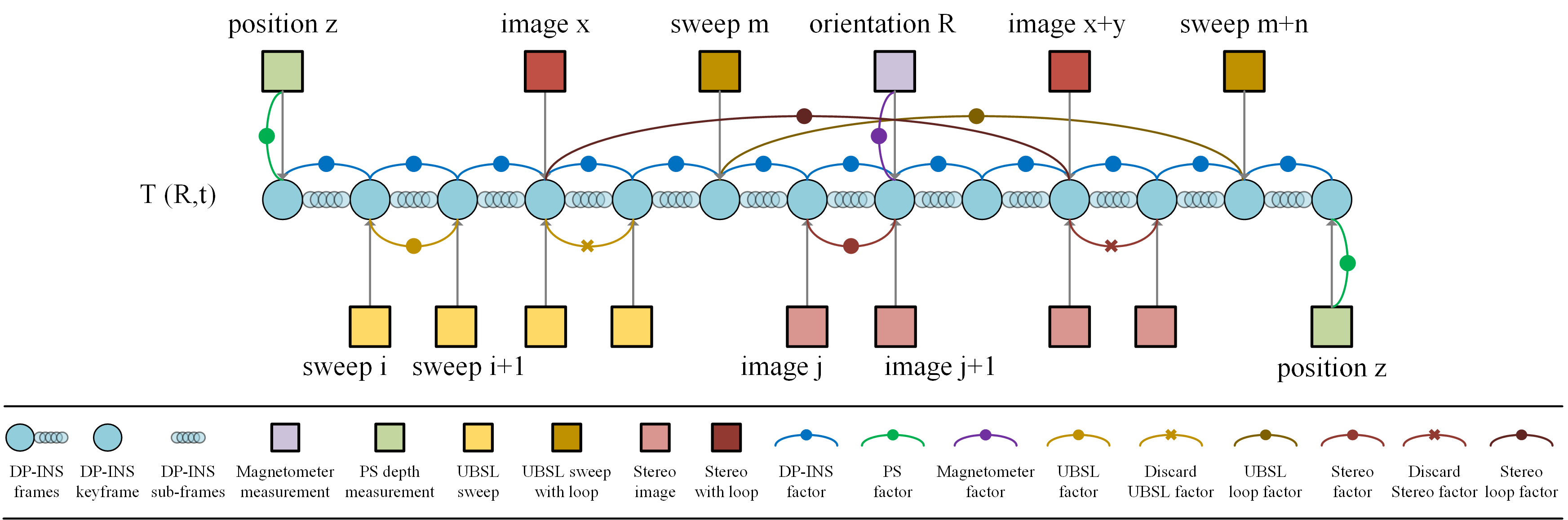}
\caption{{The schematic of multi-modal factor graph in back-end of Water-DSLAM.}}
\label{fig:factor-graph}
\end{figure*}
\subsubsection{DP-INS Factor}
In sliding windows, suppose that the current keyframe ${k_i}$'s pose is  ${\boldsymbol{T}}^{\mathcal{I}}_{k_i}$, and the previous key frame transformation matrix is ${\boldsymbol{T}}^{\mathcal{I}}_{k_{i-1}}$. 
Similarly, assuming that the DP-INS odometry attitude matrices corresponding to frames $k_i$ and $k_{i-1}$ are ${\boldsymbol{T}}^{\mathcal{I}}_{{d}_i}$ and ${\boldsymbol{T}}^{\mathcal{I}}_{{d}_{i-1}}$, respectively, the current keyframe DP-INS factor is given by:
\begin{equation}
    \boldsymbol{r}_{\mathcal{INS}}(\mathcal{V}_{i},\mathcal{D}_{i},\mathcal{X}) = \log_{\mathrm{SE(3)}}((\boldsymbol{T}^{w}_{{d}_i})^{-1}  \boldsymbol{T}^{w}_{{d}_{i-1}}  ({\boldsymbol{T}}^{w}_{k_{i-1}})^{-1}  {\boldsymbol{T}}^{w}_{k_i})^\vee,
\end{equation}
where $\log_{\mathrm{SE(3)}}(\cdot)^\vee$ denotes the map from the three-dimensional rotation group SE(3) to the Lie-Algebra $\mathfrak{se}(3)$.

\subsubsection{MAP Estimation}
To estimate the robot state within the window, the Maximum A Posteriori (MAP) estimation problem is formulated as minimizing the combination of the prior term and the Mahalanobis norm of all observations:
\begin{equation}
\begin{aligned}
&\min_{\mathcal{X}}
\left\{\begin{array}{l}\|\mathbf{r}_{p}-\mathbf{H}_{p}\mathcal{X}\|^{2}+
\sum_{k\in\mathcal{W}}\|\boldsymbol{r}_{\mathcal{V}_1}({\mathcal{L}_{i}},\mathcal{R}_i,\mathcal{X})\|_{\Sigma_{k}^{\mathcal{V}_1}}^{2}\\
+\sum_{l\in\mathcal{W}}\left\|\boldsymbol{r}_{\mathcal{V}_2}({\mathcal{L}_{i}},\hat{\mathcal{D}}^{c^r_i},\mathcal{X})\right\|_{\Sigma_{l}^{V_2}}^{2}\\
+\sum_{h\in\mathcal{W}}\left\|\boldsymbol{r}_{\mathcal{INS}}(\mathcal{V}_{i},\mathcal{D}_i,\mathcal{X})\right\|_{\Sigma_{h}^{\mathcal{INS}}}^{2},
\end{array}\right.
\end{aligned}
\end{equation}
where $\mathcal{W}$ is the measurements number in the sliding window. $\mathbf{r}_{p}$, $\mathbf{H}_{p}$ represent the prior information from marginalization.

\subsubsection{Reinitialization}
Upon detecting system failures (e.g., tracking loss or pose divergence), Water-Stereo initiates a recovery protocol: (1) Halts stereo tracking and clears buffered frames; (2) Aggregates a sliding window of image frames with timestamp-interpolated DP-INS odometry; (3) Assigns frame poses using DP-INS solutions for camera extrinsics initialization; (4) Performs stereo triangulation on feature correspondences to reconstruct visual landmarks. This hybrid approach ensures metric-scale recovery without external sensors, while the DP-INS prior constrains the solution space for numerical stability during relocalization. The reprojection error of reactivated landmarks is incorporated into subsequent bundle adjustment to refine the initial estimate.

\section{Multi-Modal Information Fusion in Back-End}
In the back-end of Water-DSLAM, we propose an asynchronous multi-modal fusion approach based on factor graph optimization, where the hard-earned relative constraints from Water-UBSL and Water-Stereo are tightly integrated to mitigate the accumulated drift of DP-INS, thereby improving the consistency of the generated map, as shown in Fig. \ref{fig:factor-graph}. The system comprises three key components: subsystem fault detection, factor graph definition, and asynchronous graph maintenance strategy.

\subsection{Subsystem Fault Detection} 
\begin{figure}[!t]
\centering
\includegraphics[width=1\linewidth]{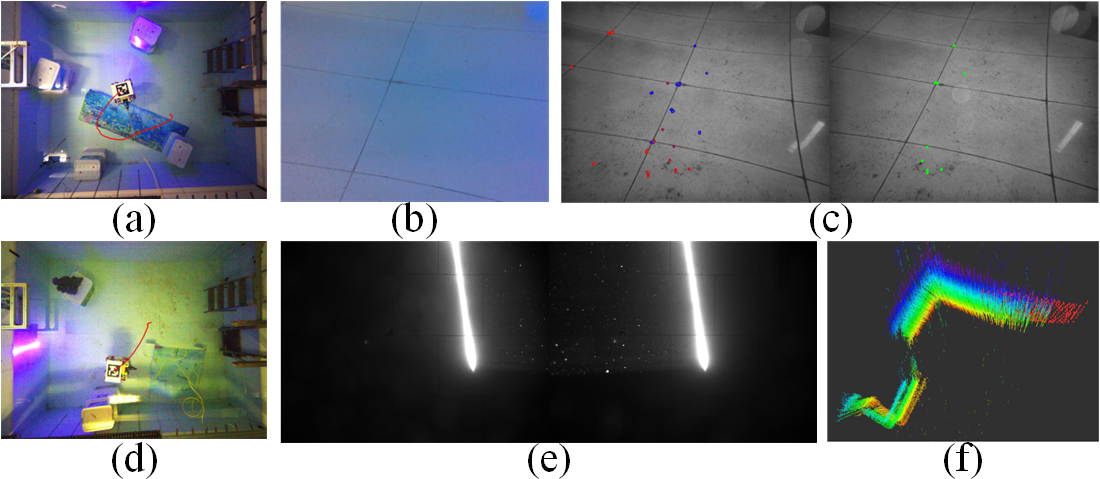}
\caption{{Subsystem fault demonstration. (a) Water-Stereo fault scenario. (b) Camera image. (c) Optical flow tracking failure. (d) Water-UBSL fault scenario. (e) UBSL image. (f) Sweep matching failure.}}
\label{fig:fault_demo}
\end{figure}

In challenging underwater scenarios, the relative constraints from Water-UBSL and Water-Stereo may degrade in textureless or structurally degenerate regions as shown in Fig. \ref{fig:fault_demo}, potentially destabilizing back-end optimization. To ensure robust performance, we design dedicated fault detection mechanisms for each subsystem to discard unreliable constraints.

\subsubsection{Fault Detection of Water-UBSL}
For the Water-UBSL, evaluating the consistency between point cloud matching results and DP-INS relative transformations.
First, the discrepancy between point cloud matching and DP-INS estimates is quantified through translational and rotational differences:
\begin{equation}
    \Delta\mathbf{t} = \|\mathbf{t}_\mathrm{UBSL} - \mathbf{t}_\mathrm{INS}\|_2, 
    \Delta\mathbf{R} = \|\log_{\mathrm{SO(3)}}(\mathbf{R}_\mathrm{UBSL}^\top \mathbf{R}_\mathrm{INS})^\vee\|_2,
\end{equation}
where $\mathbf{t}_\mathrm{UBSL},\mathbf{R}_\mathrm{UBSL}$ denote the transformation from point cloud registration, and $\mathbf{t}_\mathrm{INS},\mathbf{R}_\mathrm{INS}$ represent DP-INS relative pose estimates.
Then, to identify geometrically unreliable matches, we analyze the point cloud structure:
\begin{equation}
    \mathcal{S}_{} = 
    \begin{cases}
        1, & \text{if } N_{matched} < N_{thresh} \text{ or } \lambda_3/\lambda_1 < \epsilon\\
        0, & \text{otherwise}
    \end{cases},
\end{equation}
where $N_{thresh}$ defines the minimum required matched points, and $\lambda_i$ are eigenvalues from principal component analysis of the matched point cloud with $\epsilon$ as the planarity threshold.
A constraint is flagged as faulty if any of the following conditions are met:
\begin{equation}
    \mathcal{F} = 
    \begin{cases}
        1, & \text{if } (\Delta\mathbf{t} > \tau_t) \lor (\Delta\mathbf{R} > \tau_R) \lor (\mathcal{S}_{} = 1)\\
        0, & \text{otherwise}
    \end{cases},
\label{eq:ubsl_fault}
\end{equation}
The adaptive thresholds $\tau_t = 2.0\sigma_t^\mathrm{INS}$ and $\tau_R = 4.0\sigma_R^\mathrm{INS}$ account for DP-INS measurement uncertainty.

\subsubsection{Fault Detection of Water-Stereo}
The Water-Stereo employs the fault detection by comparing optical-flow-based pose estimation with DP-INS measurements. 
First, for consecutive stereo frames at timestamps $k$ and $k+1$, compute the relative pose residual:
\begin{equation}
\Delta\mathbf{T} = \mathbf{T}_\mathrm{Stereo}^{k+1,k} \ominus \mathbf{T}_\mathrm{INS}^{k+1,k} \in \mathfrak{se}(3).
\end{equation}
Decompose it into translational and rotational components:
\begin{equation}
\mathbf{r} = [\mathbf{r}_t, \mathbf{r}_R]^\top = \left[\|\Delta\mathbf{t}\|_2,\ \|\log_{\mathrm{SO(3)}}(\Delta\mathbf{R})^\vee\|_2\right]^\top.
\end{equation}
Then assess the visual quality by:
\begin{equation}
q_{track} = 
\underbrace{\left(\frac{N_{tracked}}{N_{total}}\right)}_{\text{Tracking rate}} \cdot 
\underbrace{\left(1 - \frac{\sigma_{flow}}{\lambda_{max}}\right)}_{\text{Flow consistency}}, \quad q_{track} \in [0,1]
\end{equation}
where $\lambda_{max}$ denotes the maximum optical flow displacement.
So the fault decision criteria is:
\begin{equation}
\mathcal{F} = 
\begin{cases}
1, & \text{if } q_{track} < q_{\mathrm{thresh}} \text{ or } \mathbf{r}_t > \tau_t \text{ or } \mathbf{r}_R > \tau_R \\
0, & \text{otherwise}
\end{cases},
\end{equation}
where $q_{\mathrm{thresh}}$ is the minimum visual quality level. $\tau_t,\tau_R$ are defined similarly to those in Eq. \ref{eq:ubsl_fault}.

\subsection{Factor Graph Definition}
\subsubsection{Optimized Node}
To compensate for the accumulated drift introduced by front-end subsystems, the back-end optimization is performed over the robot's poses. Thus, each optimized node $\boldsymbol{y}_i$ is defined as:
\begin{equation}
\boldsymbol{y}_i=\left[\begin{array}{llllllll}
\boldsymbol{p}_i^T &  \boldsymbol{R}_i^T 
\end{array}\right]{,}
\end{equation}
where $\boldsymbol{p}_i$ represents the position and $\boldsymbol{R}_i$ represents the orientation of the robot.
The DP-INS subsystem, driven by internally stable sensors, serves as the backbone for generating factor graph nodes, ensuring continuous operation of the Water-DSLAM framework. To improve efficiency and reduce redundancy, its absolute odometry at 5 Hz is used as initial values for new nodes. 
A sliding window optimization is then applied over multi-frame state variables, formulated as:
\begin{equation}
    {\mathcal{Y}} = \left[\begin{array}{lllll}
\boldsymbol{y}_1 & \boldsymbol{y}_2 & \boldsymbol{y}_3 &,...,& \boldsymbol{y}_n
\end{array}\right]{,}
\end{equation}
where $n$ indicates the window size.
As shown in Fig. \ref{fig:factor-graph}, constraint factors from three subsystems are incorporated into the graph to optimize these state variables.

\subsubsection{DP-INS Factor}
The DP-INS factor is constructed to minimize the discrepancy between the estimated poses from the factor graph and the odometry measurements provided by the DP-INS subsystem.
Suppose that the current node ${\boldsymbol{y}_i}$'s pose is  ${\boldsymbol{T}}^{\mathcal{I}}_{\boldsymbol{y}_i}$, and the previous ndoe transformation matrix is ${\boldsymbol{T}}^{\mathcal{I}}_{\boldsymbol{y}_{i-1}}$. 
Similarly, assuming that the DP-INS odometry attitude matrices corresponding to node ${\boldsymbol{y}_i}$ and ${\boldsymbol{y}_{i-1}}$ are ${\boldsymbol{T}}^{\mathcal{I}}_{{d}_i}$ and ${\boldsymbol{T}}^{\mathcal{I}}_{{d}_{i-1}}$, respectively.
The DP-INS factor is defined as:
\begin{equation}
    \boldsymbol{r}_{\mathcal{DP-INS}}(\mathcal{Y}) = \log_{\mathrm{SE(3)}}((\boldsymbol{T}^{\mathcal{I}}_{{d}_i})^{-1}  \boldsymbol{T}^{\mathcal{I}}_{{d}_{i-1}}  ({\boldsymbol{T}}^{\mathcal{I}}_{\boldsymbol{y}_{i-1}})^{-1}  {\boldsymbol{T}}^{\mathcal{I}}_{{\boldsymbol{y}}_i})^\vee.
\end{equation}

\subsubsection{PS Factor}
In underwater SLAM, absolute observations are highly valuable due to the difficulty of acquiring reliable global measurements. While PS data is already integrated into the DP-INS subsystem, its absolute constraint in the depth direction can be further leveraged to improve back-end accuracy. For this reason, we introduced seemingly redundant stressors into the factor graph.
By converting the pressure reading to a depth value, the sensor observation is denoted as $\boldsymbol{p}_\mathrm{PS}$, and $\boldsymbol{p}_0$ represents the initial absolute depth at the starting point. The resulting PS factor is defined as:
\begin{equation}
        \boldsymbol{r}_{\mathcal{PRESSURE}}(\mathcal{Y}) = {\boldsymbol{R}_{\mathcal{I}}^{\mathcal{W}}}^\top(\boldsymbol{p}_\mathrm{PS}-\boldsymbol{p}_0)-\boldsymbol{p}_i.
\end{equation}

\subsubsection{Magnetometer Factor}
In general, a 9-axis IMU provides absolute orientation observations via the magnetometer, which is essential for mitigating long-term orientation drift. To effectively utilize this information, a magnetometer factor is introduced into the factor graph framework.
The magnetometer factor is defined as:
\begin{equation}
\boldsymbol{r}_{\mathcal{MAGNETOMETER}}(\mathcal{Y})=\log_{\mathrm{SO(3)}}\left(\boldsymbol{R}_{i}^\top \boldsymbol{R}^{\mathcal{I}}_{\mathcal{M}}\boldsymbol{R}_\mathrm{Mag}\right)^\vee.
\end{equation}

\subsubsection{UBSL Factor}
For UBSL factor, suppose that the current node ${\boldsymbol{y}_i}$'s pose is  ${\boldsymbol{T}}^{\mathcal{I}}_{\boldsymbol{y}_i}$, and the previous ndoe transformation matrix is ${\boldsymbol{T}}^{\mathcal{I}}_{\boldsymbol{y}_{j}}$. 
The  transformation matrix between the current node ${\boldsymbol{y}_j}$ and the previous node ${\boldsymbol{y}_j}$ from the Water-UBSL subsystem is $\hat{\boldsymbol{T}}^{\boldsymbol{y}_i}_{\boldsymbol{y}_j}$. 
So the UBSL factor is defined as:
\begin{equation}
    \boldsymbol{r}_{\mathcal{UBSL}}\left(\mathcal{Y}\right) = \log_{\mathrm{SE(3)}}(\hat{\boldsymbol{T}}^{\boldsymbol{y}_i}_{\boldsymbol{y}_j}  ({\boldsymbol{T}}^{\mathcal{W}}_{\boldsymbol{y}_j})^{-1}  {\boldsymbol{T}}^{\mathcal{W}}_{\boldsymbol{y}_i})^\vee.
\end{equation}
\subsubsection{Stereo Factor}
For Stereo factor, suppose that the current node ${\boldsymbol{y}_i}$'s pose is  ${\boldsymbol{T}}^{\mathcal{I}}_{\boldsymbol{y}_i}$, and the previous ndoe transformation matrix is ${\boldsymbol{T}}^{\mathcal{I}}_{\boldsymbol{y}_{j}}$. 
The  transformation matrix between the current node ${\boldsymbol{y}_j}$ and the previous node ${\boldsymbol{y}_j}$ from the Water-Stereo subsystem is $\hat{\boldsymbol{T}}^{\boldsymbol{y}_i}_{\boldsymbol{y}_j}$. 
So the Stereo factor is defined as:
\begin{equation}
    \boldsymbol{r}_{\mathcal{STEREO}}\left(\mathcal{Y}\right) = \log_{\mathrm{SE(3)}}(\hat{\boldsymbol{T}}^{\boldsymbol{y}_i}_{\boldsymbol{y}_j}  ({\boldsymbol{T}}^{\mathcal{W}}_{\boldsymbol{y}_j})^{-1}  {\boldsymbol{T}}^{\mathcal{W}}_{\boldsymbol{y}_i})^\vee.
\end{equation}

\subsubsection{Loop Factor}
Building upon \cite{ou2024hybrid}, Water-DSLAM integrates information from the DP-INS, Water-Stereo, and Water-UBSL subsystems to achieve accurate and reliable loop closure detection.
Suppose that the current node ${\boldsymbol{y}_j}$ is identified as being looped back from node ${\boldsymbol{y}_i}$ by the loop closure detection algorithm, and the loop transformation matrix is $\hat{\boldsymbol{T}}^{\boldsymbol{y}_i}_{\boldsymbol{y}_j}$. 
Similarly, assuming that the odometry attitude matrices corresponding to nodes $\boldsymbol{y}_i$ and $\boldsymbol{y}_j$ are ${\boldsymbol{T}}^{\mathcal{I}}_{\boldsymbol{y}_i}$ and ${\boldsymbol{T}}^{\mathcal{I}}_{\boldsymbol{y}_j}$, respectively, the loop factor is defined as:
\begin{equation}
    \boldsymbol{r}_{\mathcal{LOOP}}\left(\mathcal{Y}\right) = \log_{\mathrm{SE(3)}}(\hat{\boldsymbol{T}}^{\boldsymbol{y}_i}_{\boldsymbol{y}_j}  ({\boldsymbol{T}}^{\mathcal{I}}_{\boldsymbol{y}_j})^{-1}  {\boldsymbol{T}}^{\mathcal{I}}_{\boldsymbol{y}_i})^\vee.
\end{equation}

\subsubsection{MAP Estimation}
By integrating all factors, MAP estimation over the entire set of nodes $\mathcal{Y}$ is performed to obtain the optimized poses for mapping:
\begin{equation}
\begin{aligned}
&\min_{\mathcal{X}}
\left\{\begin{array}{l}\sum_{\boldsymbol{x}_i\in\mathcal{Y}}\left\|\boldsymbol{r}_{\mathcal{DP-INS}}\left(\mathcal{Y}\right)\right\|_{\Sigma_{\mathrm{DP-INS}}^{\mathcal{Y}}}^{2}\\
+\sum_{\boldsymbol{x}_i\in\mathcal{Y}}\left\|\boldsymbol{r}_{\mathcal{PRESSURE}}\left(\mathcal{Y}\right)\right\|_{\Sigma_{\mathrm{pressure}}^{\mathcal{Y}}}^{2}\\
+\sum_{\boldsymbol{x}_i\in\mathcal{Y}}\left\|\boldsymbol{r}_{\mathcal{MAGNETOMETER}}\left(\mathcal{Y}\right)\right\|_{\Sigma_{\mathrm{mag}}^{\mathcal{Y}}}^{2}\\
+\sum_{\boldsymbol{x}_i\in\mathcal{Y}}\left\|\boldsymbol{r}_{\mathcal{UBSL}}\left(\mathcal{Y}\right)\right\|_{\Sigma_{\mathrm{UBSL}}^{\mathcal{Y}}}^{2}\\
+\sum_{\boldsymbol{x}_i\in\mathcal{Y}}\left\|\boldsymbol{r}_{\mathcal{STEREO}}\left(\mathcal{Y}\right)\right\|_{\Sigma_{\mathrm{stereo}}^{\mathcal{Y}}}^{2}\\
+\sum_{\boldsymbol{x}_i\in\mathcal{Y}}\left\|\boldsymbol{r}_{\mathcal{LOOP}}\left(\mathcal{Y}\right)\right\|_{\Sigma_{\mathrm{loop}}^{\mathcal{Y}}}^{2}.
\end{array}\right.
\end{aligned}
\end{equation}

\subsection{Asynchronous Graph Maintenance Strategy}
\begin{figure}[!t]
\centering
\includegraphics[width=1\linewidth]{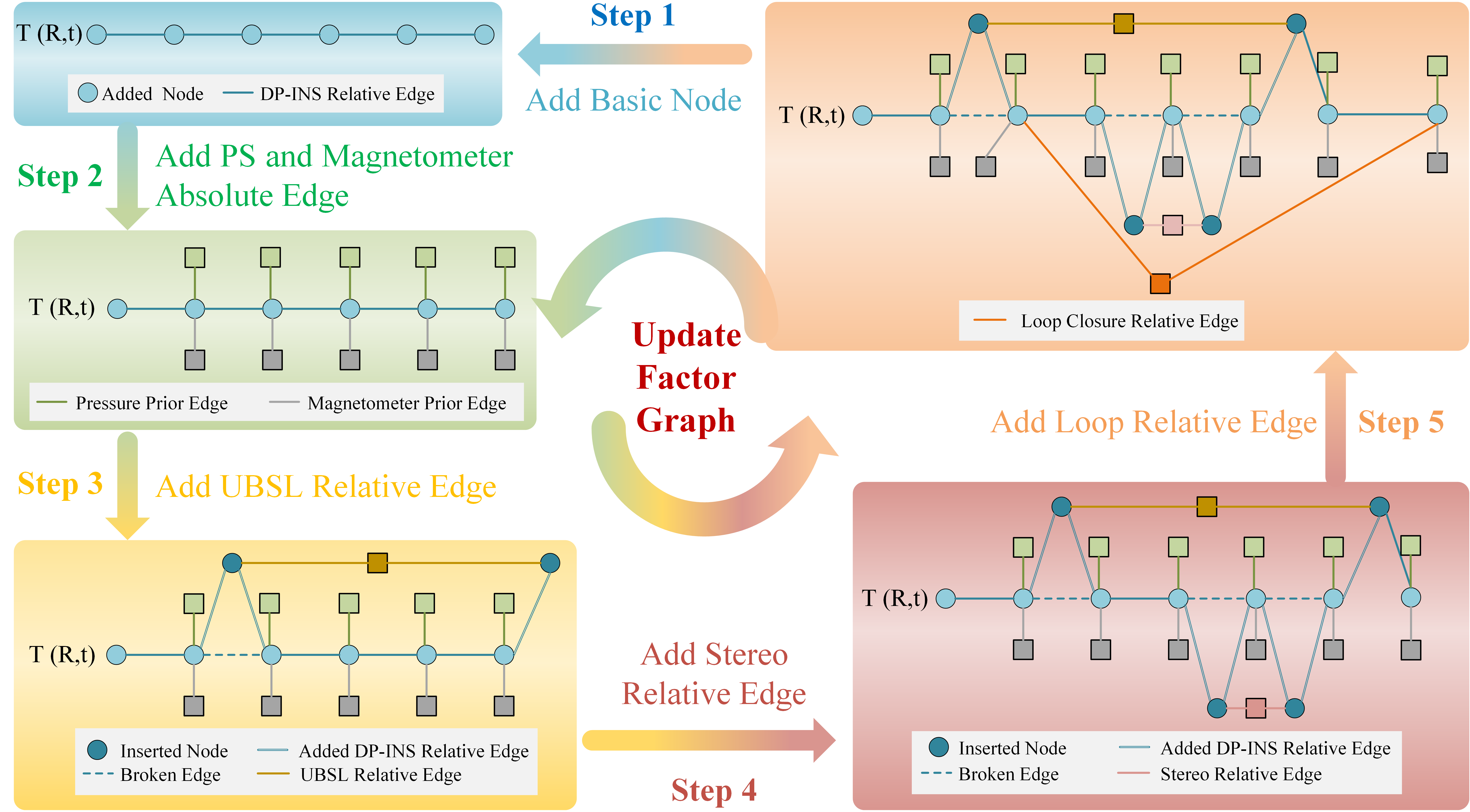}
\caption{{The maintenance process of multi-modal factor graph.}}
\label{fig:node update}
\end{figure}

In challenging underwater environments, external sensors are prone to intermittent failure, resulting in asynchronously and sporadically arriving factors. To address this unpredictability, we propose a flexible, fault‑tolerant factor‑graph maintenance algorithm that accommodates arbitrarily arriving heterogeneous factors. This method effectively mitigates the adverse effects of asynchronous sensor rates, intermittent data streams, and partial sensor drop‑outs, ensuring robust graph‑based SLAM under uncertain sensing conditions. The cycle steps are shown in Fig. \ref{fig:node update}.

\subsubsection{Step 1: Add New Node and DP-INS Absolute Edge}
In Water-DSLAM, DP-INS provides high-frequency and relatively stable odometry, making it suitable for node creation. After downsampling to \~5 Hz, the factor graph uses DP-INS data to initialize node states with prior edges, followed by the addition of absolute edges from DP-INS.

\subsubsection{Step 2: Add PS and Magnetometer Absolute Edge}
In the second step, the absolute edges provided by the high-frequency PS and magnetometer data buffer are interpolated for each node created in Step 1. 

The data from the first two steps are relatively stable and require minimal handling of anomalies. In contrast, the Water-UBSL, Water-Stereo subsystems, and loop closure module are subject to significant uncertainty due to the complex underwater environment, with both valid and invalid data exhibiting randomness. Moreover, these factors operate at lower frequencies and involve interactions with historical nodes, making node addition and removal more challenging. To maintain consistency, we represent all three factors in the unified form of $(t_1, t_2, \boldsymbol{T})$.
\subsubsection{Step 3: Add UBSL Relative Edge}
For UBSL data $(t_1, t_2, \boldsymbol{T}_{\mathrm{UBSL}})$, we first store the corresponding sweep for back-end mapping. To process $t_1$, we identify the adjacent historical nodes $n_1$ and $n_2$ with timestamps $t_{n_1}$ and $t_{n_2}$, insert a new node $n_{t_1}$, and retrieve $\boldsymbol{T}_{\mathrm{DP-INS},t_1}$ from the DP-INS buffer. 
Two DP-INS relative edges are then added: $n_1 \rightarrow n_{t_1}$ and $n_{t_1} \rightarrow n_2$.
For $t_2$, a new node $n_{t_2}$ is appended after the current end node $n_{end}$, with $\boldsymbol{T}_{\mathrm{DP-INS},t_2}$ interpolated from the buffer. A DP-INS relative edge is added between $n_{end}$ and $n_{t_2}$. Finally, a UBSL relative edge based on $T_{\mathrm{UBSL}}$ is inserted between $n_{t_1}$ and $n_{t_2}$.

\subsubsection{Step 4: Add Stereo Relative Edge}
For Stereo data $(t_1, t_2, \boldsymbol{T}_{\mathrm{Stereo}})$, we first store the associated sweep for backend mapping. Similar to Step 3, we locate adjacent nodes $n_1$ and $n_2$ around $t_1$, insert node $n_{t_1}$, and retrieve $T_{\mathrm{DP-INS},t_1}$ to add two DP-INS relative edges: $n_1 \rightarrow n_{t_1}$ and $n_{t_1} \rightarrow n_2$.
At $t_2$, we insert node $n_{t_2}$ after $n_{end}$, interpolate $T_{\mathrm{DP-INS},t_2}$, and add a DP-INS edge between $n_{end}$ and $n_{t_2}$. Finally, a Stereo relative edge is added between $n_{t_1}$ and $n_{t_2}$ using $\boldsymbol{T}_{\mathrm{Stereo}}$.

\subsubsection{Step 5: Add Loop Relative Edge}
For loop closure data $(t_1, t_2, \boldsymbol{T}_{\mathrm{Loop}})$, we store the corresponding sweep and directly locate historical nodes $n_{t_1}$ and $n_{t_2}$. A loop relative edge is then inserted between them based on $\boldsymbol{T}_{\mathrm{Loop}}$.

\subsection{Dense Mapping}
Mapping is conducted on a per-sweep basis. During UBSL factor construction, sweeps from the Water-UBSL subsystem are synchronously received. After several optimization iterations, the mapping process begins. The poses corresponding to each sweep’s timestamp are retrieved from the factor graph and interpolated, allowing the sweep point clouds to be transformed into a unified coordinate frame. So Water-Scanner realizes the reconstruction of a rarely dense point cloud map for underwater in-situ observation.

\section{Experiments}

\subsection{Experiments Setup}
\begin{figure}[!t]
\centering
\includegraphics[width=1\linewidth]{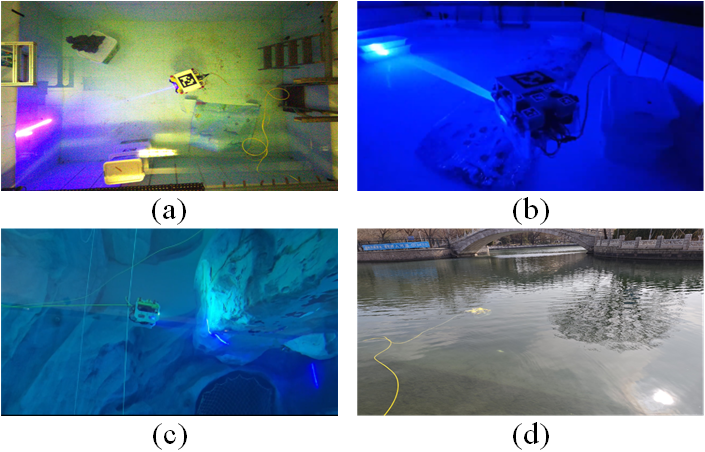}
\caption{{Experimental scenes. (a) Pool. (b) Dark underwater. (c) 16-meter-depth sinkhole. (d) Field river.}}
\label{fig:exp_scene}
\end{figure}
To thoroughly validate the superior performance of the proposed framework Water-DSLAM, we used the designed Water-Scanner to collect data in various complex underwater scenarios, including pools, dark underwater scenes, 16-meter-deep sinkholes, and field rivers, as shown in Fig. \ref{fig:exp_scene}.

Specifically, ablation experiments of Water-DSLAM were conducted in a real pool environment to comprehensively demonstrate the importance of each subsystem for underwater applications. Subsequently, self-stabilizing localized fine-scale inspection experiments were performed to verify the advantages of Water-DSLAM in detailed in-situ underwater observation. Furthermore, random free-motion mapping experiments were carried out in the pool, where Water-DSLAM was compared against current state-of-the-art (SOTA) underwater SLAM methods, highlighting its significant improvements in localization accuracy, mapping granularity, and consistency.
To further demonstrate its applicability in underwater engineering, we conducted three application experiments in more challenging conditions: (1) fine-scale mapping in dark underwater environments, (2) a 10-minute 750 $m^3$ large-scale inspection in a 16-meter-deep underwater sinkhole, and (3) in-situ observation in a field flowing river setting.

For obtaining Ground Truth trajectories, an Apriltag marker attached above the robot was used during the pool experiments as shown in Fig. \ref{fig:exp_scene}(a). This marker extended above the water surface and was detected by a global camera positioned above the pool. 
To quantify localization accuracy, EVO \cite{rebecq2016evo} was utilized for trajectory evaluation. By calculating the Root Mean Square Error (RMSE) and Mean Absolute Error (MAE) of Absolute Trajectory Error (ATE) and Relative Trajectory Error (RTE), a systematic analysis of the algorithm's performance was conducted from the perspectives of both position and orientation.
All experiments were conducted on a laptop equipped with an Intel\textsuperscript{\textregistered} Core\texttrademark{} i9-13900HX processor (2.2\,GHz base frequency, 5.4\,GHz Turbo).
Note: 
Due to limited open-source underwater research, method comparisons only reproduce core ideas on our platform.

\subsection{Subsystem Ablation Experimental Verification}
To comprehensively evaluate Water-DSLAM, a series of ablation studies were conducted. Each core subsystem—DP-INS, Water-Stereo, and Water-UBSL—was independently assessed to demonstrate its effectiveness in underwater conditions. A full-system ablation was then performed to analyze the individual contributions to the overall system performance.

\begin{figure}[!t]
\centering
\includegraphics[width=1\linewidth]{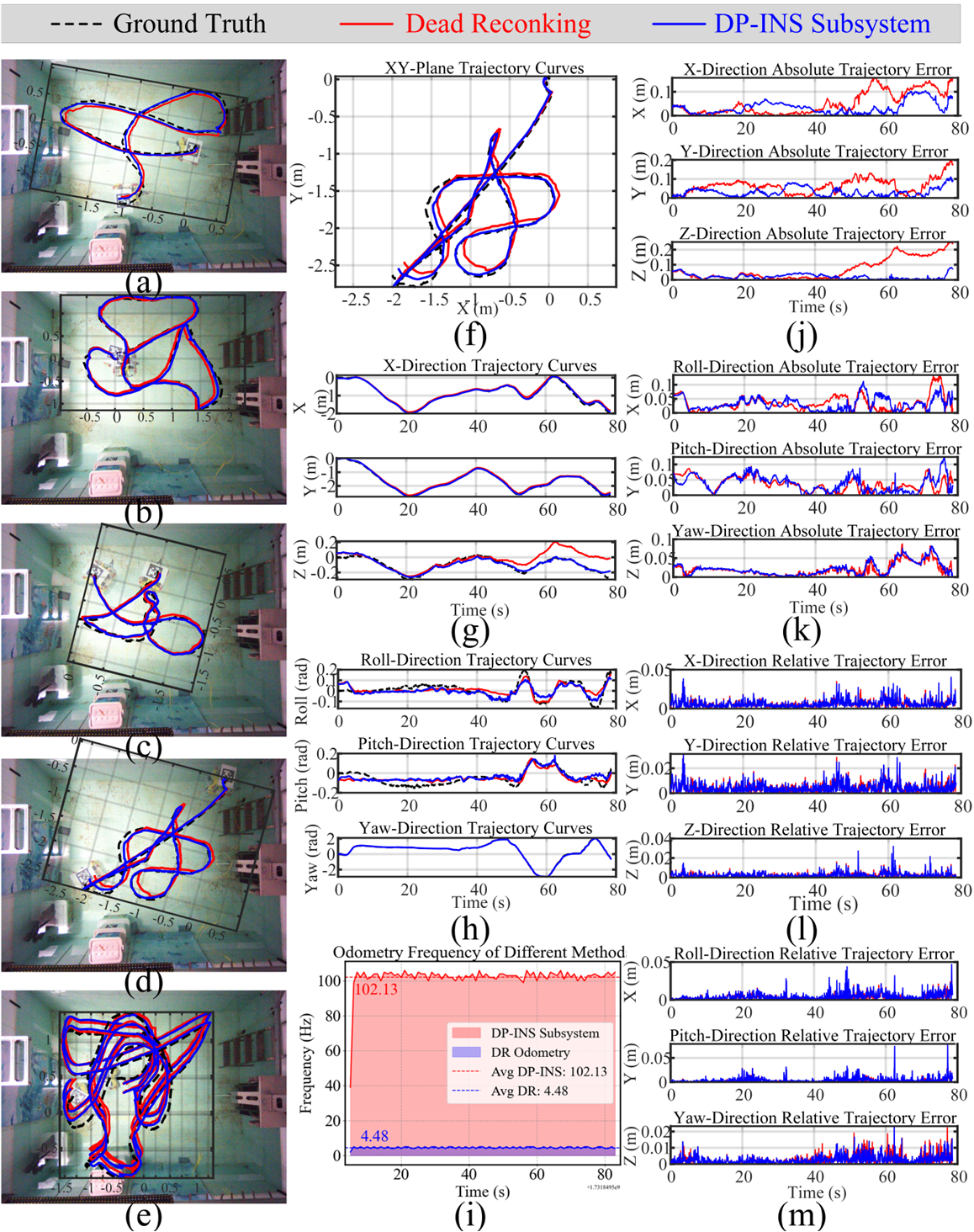}
\caption{{Experimental results of the ablation study on the DP-INS subsystem. (a)-(e) Localization performance across multiple datasets. (f)-(h) Trajectory curves in different directions in dataset (d). (i) Comparison of frequency curves in dataset (d). (j)-(m) Error curves in different directions in dataset (d).
}}
\label{fig:exp_b_1}
\end{figure}
\begin{table*}[!htbp]
\label{tab:pool_dp-ins}
{\caption{Quantitative Results of Ablation Study on the DP-INS Subsystem.\label{tab:pool_dp-ins}}}
\centering
\begin{tabular}{c|@{\hspace{5pt}}c|@{\hspace{5pt}}c|@{\hspace{5pt}}c|@{\hspace{5pt}}c||c@{\hspace{5pt}}c|c@{\hspace{5pt}}c|c@{\hspace{5.5pt}}c@{\hspace{5.5pt}}|c@{\hspace{5.5pt}}c@{\hspace{5pt}}|c}
\Xhline{1.2pt}
\multirow{2}{*}{Datasets}&\multirow{2}{*}{Dura. (s)}& \multirow{2}{*}{Path (m)}& \multirow{2}{*}{\begin{tabular}[c]{@{}c@{}}Ave. Vel.\\(m/s)\end{tabular}}&\multirow{2}{*}{Method}& \multicolumn{2}{c|}{ATE Pos. (m)}   & \multicolumn{2}{c|}{ATE Rot. (deg)}  & \multicolumn{2}{c|}{RTE Pos. (m)} & \multicolumn{2}{c|}{RTE Rot. (deg)} &\multirow{2}{*}{\begin{tabular}[c]{@{}c@{}}Frequency \\ (Hz)\end{tabular}}\\ 
     \cmidrule[0.5pt](rl){6-7}
    \cmidrule[0.5pt](rl){8-9}
    \cmidrule[0.5pt](rl){10-11}
    \cmidrule[0.5pt](rl){12-13}
& & & & &RMSE  & MAE  & RMSE   & MAE  &  RMSE    & MAE   &  RMSE  & MAE & \\ \Xhline{1.2pt}
\multirow{2}{*}{{\textbf{\#PoolAbla1}}}&\multirow{2}{*}{{56.7}}&\multirow{2}{*}{{8.8}}&\multirow{2}{*}{{0.155}} &DR \cite{huang2024visual}&{0.098}  &  \textbf{0.080}   & 4.896 & 4.502 & 0.023  & 0.018  & 1.054  & 0.821 & 4.47\\
& & & & DP-INS& \textbf{0.093}  & 0.082  & \textbf{3.516}  &  \textbf{3.216}   & \textbf{0.009}     & \textbf{0.006} & \textbf{0.462}   & \textbf{0.351}  &$\textbf{100.86}_{(\blue{{\textbf{22.6}\times^\uparrow}})}$ \\ \hline
\multirow{2}{*}{\textbf{\#PoolAbla2}}&\multirow{2}{*}{{58.5}}&\multirow{2}{*}{{8.4}}&\multirow{2}{*}{{0.154}} &DR \cite{huang2024visual}&\textbf{0.057}  & \textbf{0.050}   & 6.665 & 5.977 & 0.023  & 0.018  & 1.414  & 1.190 & 4.46\\
& & & & DP-INS & 0.058  & \textbf{0.050}  & \textbf{3.775}  &  \textbf{3.463}   & \textbf{0.008}     & \textbf{0.006} & \textbf{0.714}   & \textbf{0.538}  &$\textbf{101.27}_{(\blue{{\textbf{22.7}\times^\uparrow}})} $\\\hline
\multirow{2}{*}{\textbf{\#PoolAbla3}}&\multirow{2}{*}{{78.6}}&\multirow{2}{*}{{13.6}}&\multirow{2}{*}{{0.174}} &DR \cite{huang2024visual}&0.092  & 0.080   & 5.425 & 5.197 & 0.025  & 0.020  & 1.307  & 0.922 & 4.48\\
& & & & DP-INS & \textbf{0.076}  & \textbf{0.066}  & \textbf{3.797}  & \textbf{3.542}   & \textbf{0.008}     & \textbf{0.007} & \textbf{0.603}   & \textbf{0.439}  & $\textbf{102.13}_{(\blue{{\textbf{22.8}\times^\uparrow}})}$\\\hline
\multirow{2}{*}{\textbf{\#PoolAbla4}}&\multirow{2}{*}{{78.2}}&\multirow{2}{*}{{13.3}}&\multirow{2}{*}{{0.170}} &DR \cite{huang2024visual}&0.084  &  \textbf{0.075}   & 9.568 & 9.249 & 0.024  & 0.020  & 1.532  & 1.132 & 4.41\\
& & & & DP-INS & \textbf{0.081}  & \textbf{0.075}  & \textbf{4.113}  &  \textbf{3.821}   & \textbf{0.008}     & \textbf{0.006} & \textbf{0.669}   & \textbf{0.488}  &$\textbf{101.59}_{(\blue{{\textbf{23.0}\times^\uparrow}})} $\\\hline
\multirow{2}{*}{\textbf{\#PoolAbla5}}&\multirow{2}{*}{{98.5}}&\multirow{2}{*}{{18.4}}&\multirow{2}{*}{{0.188}} &DR \cite{huang2024visual}&0.122  & 0.106   & 5.924 &5.429 & 0.031  & 0.024  & 1.990  & 1.457 &4.44 \\
& & & & DP-INS & \textbf{0.102}  & \textbf{0.089}  & \textbf{4.261}  &  \textbf{3.844}   & \textbf{0.010}     & \textbf{0.007} & \textbf{0.695}   & \textbf{0.489}  & $\textbf{101.75}_{(\blue{{\textbf{22.9}\times^\uparrow}})}$\\
\Xhline{1.2pt}
\end{tabular}
\end{table*}
\subsubsection{Ablation Study of DP-INS Subsystem}
Compared to the low-frequency trajectory data typically provided by conventional DR systems~\cite{huang2024visual}, we propose a DP-INS subsystem that integrates IMU, DVL, and PS to generate higher-frequency odometry with more stable depth estimates. To evaluate its performance, five ablation experiments (\textbf{\#PoolAbla1} - \textbf{\#PoolAbla5} datasets) with different trajectories were conducted in a pool. As shown in Fig.~\ref{fig:exp_b_1}(a)–(e), the Water-Scanner performed random motions lasting up to 100 s, with trajectory lengths reaching 18.4 m. Details are provided in Table~\ref{tab:pool_dp-ins}. 
Results from the \textbf{\#PoolAbla4} include trajectory curves along the x-y plane, x, y, and z axes, roll, pitch, yaw, and ATE/RTE metrics, as shown in Fig.~\ref{fig:exp_b_1}(f)-(h), and (j)-(m). The DP-INS trajectory aligns more closely with the ground truth, showing significantly lower ATE and RTE than DR, especially along the z-axis. This demonstrates the advantage of multi-sensor fusion in underwater localization. As shown in Fig.~\ref{fig:exp_b_1}(i), DP-INS achieves a high-frequency odometry output of up to 100~Hz, meeting the requirements of structured light scanning, while DR remains below 5~Hz and is insufficient for dense mapping.

Furthermore, a quantitative evaluation was performed by calculating the RMSE and MAE of ATE and RTE, as summarized in Table \ref{tab:pool_dp-ins}. The results show that DP-INS effectively reduces both ATE and RTE, alleviating the cumulative error typically seen in traditional acoustic-based dead reckoning. On the longest trajectory dataset (\textbf{\#PoolAbla5}, 18.4 m over 98.5 s), DP-INS reduced the ATE position RMSE from 0.122 m to 0.102 m and the ATE rotation RMSE from 5.924° to 4.261°. More notably, the RTE position RMSE dropped from 0.031 m to 0.010 m, and RTE rotation from 1.990° to 0.695°. In addition, DP-INS achieved a 22.9× increase in output frequency, making it well-suited for integration with structured-light systems.

\subsubsection{Ablation Study of Water-Stereo Subsystem}
\begin{figure}[!t]
\centering
\includegraphics[width=1\linewidth]{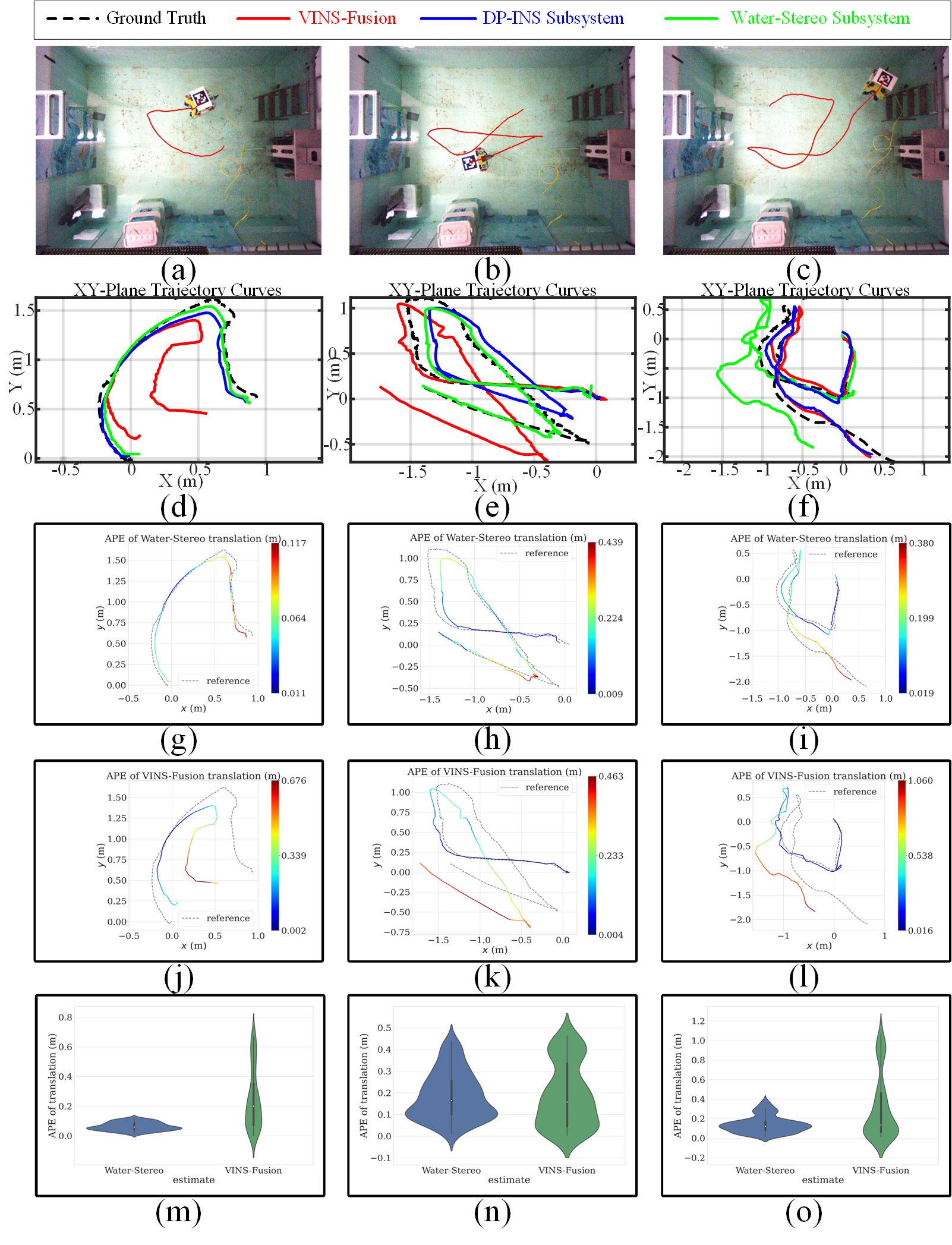}
\caption{{Experimental results of the ablation study on the Water-Stereo subsystem. (a)-(c) Demonstration of trajectories for multiple datasets. (d)-(f) XY-plane trajectory curves in different methods. (g)-(i) Visualization of the localization errors for Water-Stereo.
(j)-(l) Visualization of the localization errors for VINS-Fusion \cite{qin2019general}. (m)-(o) Violin plots of localization errors.
}}
\label{fig:exp_b_2}
\end{figure}
\begin{table}[!t]
\caption{Quantitative Results of Ablation Study on the Water-Stereo.}\label{tab:water-stereo}
\centering
\setlength{\tabcolsep}{3pt} 
\begin{tabular}{@{\hspace{0.0pt}}c|@{\hspace{0.0pt}}c|@{\hspace{0.0pt}}c||c@{\hspace{0.0pt}}c@{\hspace{0.0pt}}c@{\hspace{0.0pt}}c} 
\Xhline{1.2pt}
    \multirow{3}{*}{\begin{tabular}[c]{@{}c@{}}Datasets\end{tabular}}   &\multirow{3}{*}{\begin{tabular}[c]{@{}c@{}}Path \\(m)\end{tabular}}     &\multirow{3}{*}{\begin{tabular}[c]{@{}c@{}}Ave.\\ Vel. \\(m/s)\end{tabular}}     & \multicolumn{2}{c}{ATE RMSE (m)}  & \multicolumn{2}{c}{ATE MAE (m)}\\ 
    \cmidrule[0.5pt](rl){4-5}
    \cmidrule[0.5pt](rl){6-7}
    & & &\multirow{2}{*}{\begin{tabular}[c]{@{}c@{}}VINS-\\Fusion \cite{qin2019general}\end{tabular}}   & \multirow{2}{*}{\begin{tabular}[c]{@{}c@{}}Water-\\Stereo\end{tabular}}  & \multirow{2}{*}{\begin{tabular}[c]{@{}c@{}}VINS-\\Fusion \cite{qin2019general}\end{tabular}}   & \multirow{2}{*}{\begin{tabular}[c]{@{}c@{}}Water-\\Stereo\end{tabular}}   \\ 
    && &&&&\\
\Xhline{1.2pt}
    \textbf{\#PoolAbla6}&  3.3 &0.125 &  0.319  & $\textbf{0.068}_{(\blue{{\textbf{4.69}\times^\downarrow}})}$   & 0.248   &  $\textbf{0.063}_{(\blue{{\textbf{3.94}\times^\downarrow}})} $ \\
    \textbf{\#PoolAbla7} &6.4& 0.203 & 0.241  & $\textbf{0.215}_{(\blue{{\textbf{1.12}\times^\downarrow}})} $  & 0.192  & $\textbf{0.186}_{(\blue{{\textbf{1.03}\times^\downarrow}})} $   \\
    \textbf{\#PoolAbla8} & 7.2& 0.158 & 0.446 & $\textbf{0.165}_{(\blue{{\textbf{2.70}\times^\downarrow}})}$  & 0.305   &$\textbf{0.141}_{(\blue{{\textbf{2.16}\times^\downarrow}})} $ \\ 
\Xhline{1.2pt}
\end{tabular}
\end{table}

To enhance underwater visual SLAM robustness and accuracy, we integrated DP-INS into VINS-Fusion \cite{qin2019general} and developed the Water-Stereo subsystem, combining acoustic and visual data for more reliable pose estimation. We evaluated its performance using three datasets (\textbf{\#PoolAbla6} - \textbf{\#PoolAbla8}) collected in a feature-rich pool environment with the Water-Scanner. Fig.~\ref{fig:exp_b_2}(a)–(c) compares the trajectories of pure VINS-Fusion, pure DP-INS, and the Water-Stereo subsystem. As shown in Fig.~\ref{fig:exp_b_2}(d)–(f),the feature-rich environment ensured operational visual odometry, enabling a fair comparison. However, the VINS-Fusion exhibited significant distortion due to feature extraction errors and environmental challenges (Fig.~\ref{fig:exp_b_2}(j)–(l)). In contrast, the Water-Stereo subsystem reduced drift by incorporating acoustic measurements, offering more stable trajectory constraints (Fig.~\ref{fig:exp_b_2}(g)–(i)).

To further quantify performance, we conducted ATE-based evaluations, as shown in Fig.~\ref{fig:exp_b_2}(m)–(o). The results, summarized in Table~\ref{tab:water-stereo}, show that the Water-Stereo subsystem achieved a 2.84$\times$ reduction in RMSE and a 2.38$\times$ improvement in MAE compared to the original VINS-Fusion. These findings confirm that the integration of DP-INS significantly enhances the robustness and accuracy of underwater SLAM. Overall, the proposed Water-Stereo subsystem presents a promising solution for achieving high-precision and reliable localization in underwater environments.

\subsubsection{Ablation Study of Water-UBSL Subsystem}

\begin{figure}[!t]
\centering
\includegraphics[width=1\linewidth]{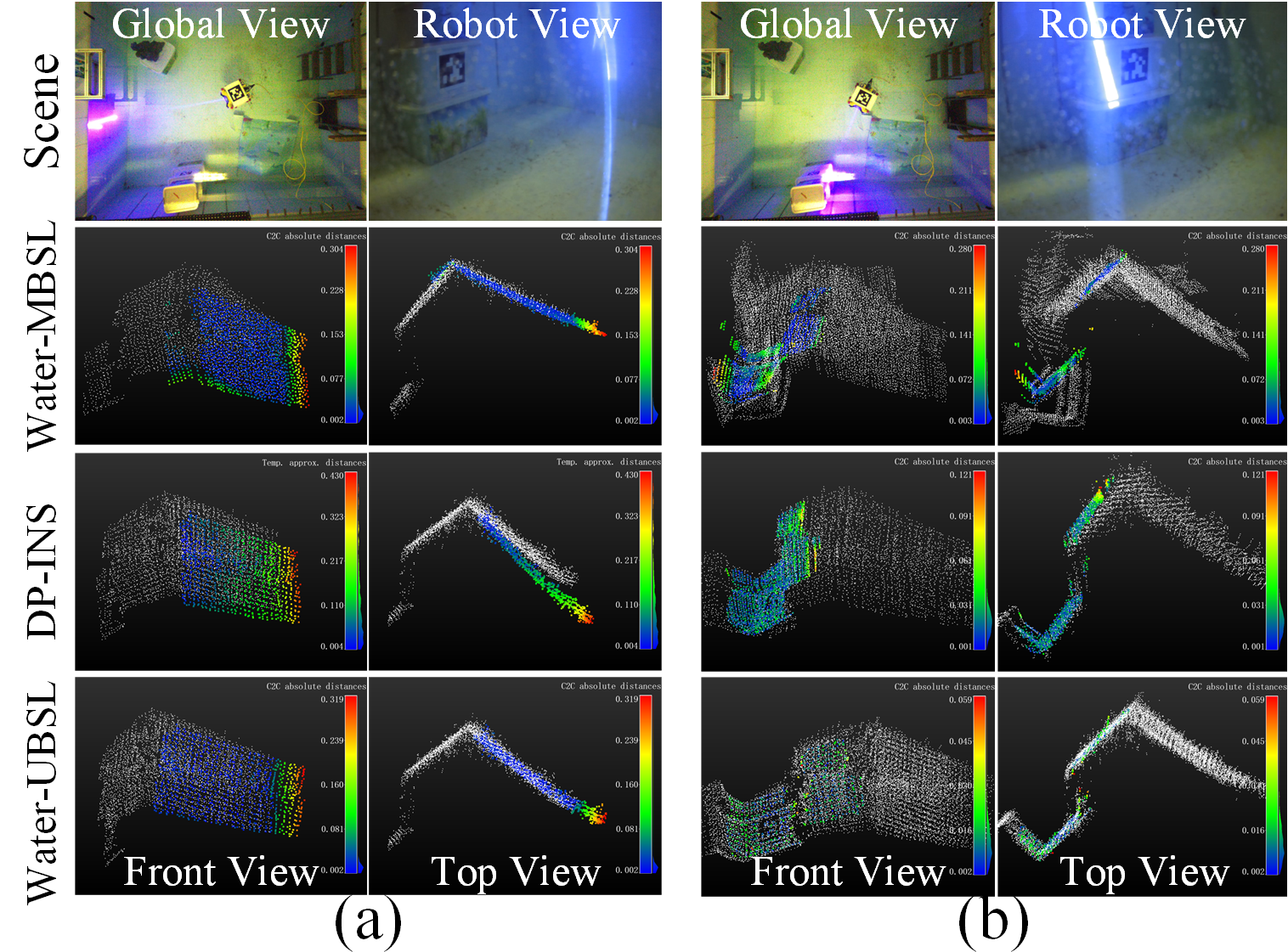}
\caption{Visualization of the point cloud matching process during the ablation study of the Water-UBSL subsystem: (a) 3rd sweep. (b) 11th sweep.}
\label{fig:b_3}
\end{figure}

To enhance mapping consistency, we incorporate UBSL information into the front-end of Water-DSLAM to provide additional relative constraints. Due to the frequent structural degradation in underwater environments, the loosely coupled Water-MBSL system based on structured light in \cite{ou2023water} often suffers from point cloud matching failures in large-scale area. To address this, we propose the Water-UBSL subsystem—a tightly coupled framework that integrates UBSL with DP-INS using the IESKF.
We validated Water-UBSL in a locally structured pool environment through ablation experiments comparing it with standalone UBSL-based Water-MBSL and DP-INS. As shown in Fig. \ref{fig:b_3}, Water-MBSL performs well in regions with distinct geometric features (e.g., Fig. \ref{fig:b_3}(a)), providing accurate pose constraints that improve mapping consistency. However, under rapid motion, the degradation of initial estimates leads to mismatches in point cloud alignment and a drop in consistency (Fig. \ref{fig:b_3}(b)). While DP-INS alone offers coarse alignment, it lacks the precision needed for fine mapping. In contrast, the tightly integrated Water-UBSL subsystem significantly improves the reliability of structured light-based matching and mitigates the effects of environmental degradation underwater.

Additional quantitative results are presented in Table \ref{tab:water-ubsl}. Compared to the loosely coupled Water-MBSL and standalone DP-INS, the proposed Water-UBSL consistently achieves superior mapping accuracy over time. For the 3rd sweep match, Water-UBSL is slightly less affected by the acoustic data degradation than Water-MBSL. As the sequence progresses, the advantage becomes more evident—by the 11th sweep, Water-UBSL reduces RMSE from 0.067~m to 0.016~m and MAE from 0.049~m to 0.014~m, marking an improvement of over 70\% in both metrics. These results highlight that the tightly coupled Water-UBSL subsystem provides more accurate and robust constraints for back-end consistent mapping, especially under long-duration and degraded conditions.
    \begin{table}[!t]
    \label{tab:water-ubsl}
    \caption{Quantitative Results of Ablation Study on the Water-UBSL.\label{tab:water-ubsl}}
    \centering
        \begin{tabular}{c|@{\hspace{5pt}}c||@{\hspace{5pt}}c@{\hspace{5pt}}c@{\hspace{5pt}}c@{\hspace{5pt}}c@{\hspace{5pt}}c@{\hspace{5pt}}c}
        \Xhline{1.2pt}
              \multirow{3}{*}{\begin{tabular}[c]{@{}c@{}}Time\\ (s)\end{tabular}}   &\multirow{3}{*}{\begin{tabular}[c]{@{}c@{}}Sweep \\Count\end{tabular}}  & \multicolumn{3}{c}{ RMSE (m)}  & \multicolumn{3}{c}{MAE (m)}\\ 
                    \cmidrule[0.5pt](rl){3-5}
                    \cmidrule[0.5pt](rl){6-8}
            & &\multirow{2}{*}{\begin{tabular}[c]{@{}c@{}}Water-\\MBSL\cite{ou2023water}\end{tabular}}   & \multirow{2}{*}{\begin{tabular}[c]{@{}c@{}}DP-\\INS\end{tabular}}  & \multirow{2}{*}{\begin{tabular}[c]{@{}c@{}}Water-\\UBSL\end{tabular}}   & \multirow{2}{*}{\begin{tabular}[c]{@{}c@{}}Water-\\MBSL\cite{ou2023water}\end{tabular}}  & \multirow{2}{*}{\begin{tabular}[c]{@{}c@{}}DP-\\INS\end{tabular}}  & \multirow{2}{*}{\begin{tabular}[c]{@{}c@{}}Water-\\UBSL\end{tabular}}  \\ 
         && &&&&&\\\Xhline{1.2pt}
           6.6&  3& \red{0.073}  & 0.117   &\blue{0.089}   & \red{0.045} &0.117&\blue{0.049} \\
         10.2 &5& 0.086  & \blue{0.061}  & \red{0.031}  & 0.086  &\blue{0.047}&\red{0.027} \\
          17.6 & 11& 0.067 & \blue{0.029}&\red{0.016}  &0.049&\blue{0.024}&\red{0.014}\\ \Xhline{1.2pt}
        \end{tabular}
        \begin{tablenotes}
    \footnotesize
    \item[1. ] 1. \red{Red} text indicates the best result. \blue{Blue} text indicates the 2nd best result.
    \end{tablenotes}
    \end{table}

\subsubsection{Subsystem Ablation Experiment}
\begin{figure*}[!htbp]
\centering
\includegraphics[width=1\linewidth]{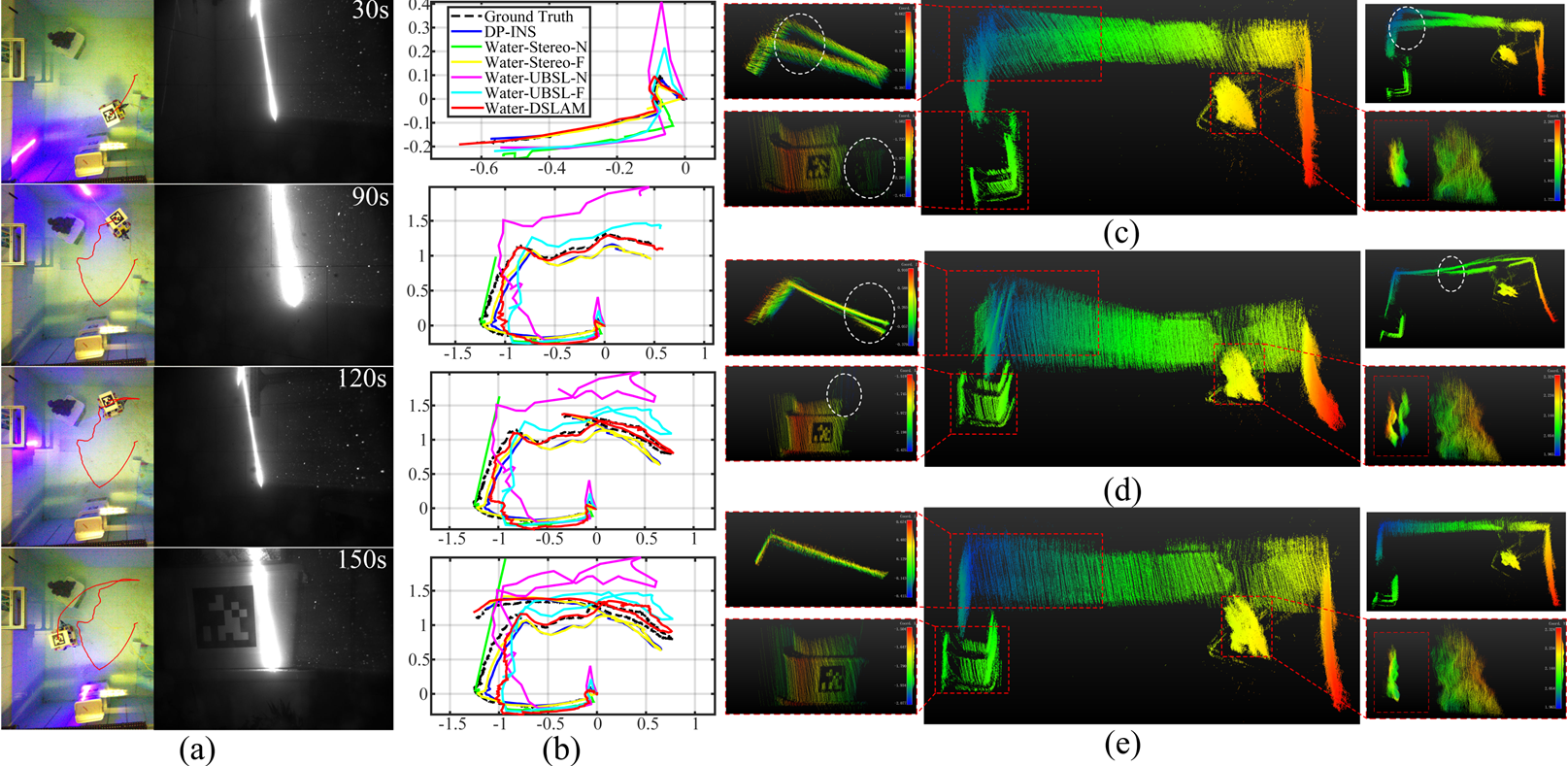}
\caption{{Experimental results of the ablation study on different subsystems of Water-DSLAM. (a) Robot movement process. (b) Trajectories at different moments. (c) Mapping results of DP-INS subsystem. (d) Mapping results of Water-UBSL subsystem. (e) Mapping results of Water-DSLAM.}}
\label{fig:b_4}
\end{figure*}
\begin{table*}[!htbp]
\label{tab:compare_pool}
{\caption{Quantitative Results of the Ablation Study on Different Subsystems of Water-DSLAM.\label{tab:compare_pool}}}
\centering
\begin{tabular}{@{\hspace{5pt}}c|@{\hspace{5pt}}c|@{\hspace{5pt}}c||c@{\hspace{5pt}}c|c@{\hspace{5pt}}c|c@{\hspace{5.5pt}}c@{\hspace{5.5pt}}|c@{\hspace{5.5pt}}c@{\hspace{5pt}}|c@{\hspace{5.5pt}}c|c@{\hspace{5.5pt}}c}
\Xhline{1.2pt}
\multirow{2}{*}{Dura. (s)}& \multirow{2}{*}{Path (m)}& \multirow{2}{*}{\begin{tabular}[c]{@{}c@{}}Ave. Vel.\\(m/s)\end{tabular}}& \multicolumn{2}{c|}{DP-INS}   & \multicolumn{2}{c|}{Water-Stereo-N}  & \multicolumn{2}{c|}{Water-Stereo-F} & \multicolumn{2}{c|}{Water-UBSL-N} &\multicolumn{2}{c|}{Water-UBSL-F}&\multicolumn{2}{c}{Water-DSLAM}\\ 
     \cmidrule[0.5pt](rl){4-5}
    \cmidrule[0.5pt](rl){6-7}
    \cmidrule[0.5pt](rl){8-9}
    \cmidrule[0.5pt](rl){10-11}
    \cmidrule[0.5pt](rl){12-13}
    \cmidrule[0.5pt](rl){14-15}
 & &  &Ratio  & ATE (m)  & Ratio   & ATE (m)  &  Ratio    & ATE (m)   &  Ratio  & ATE (m) &Ratio & ATE (m)&Ratio & ATE (m)\\ \Xhline{1.2pt}
30 & 1.6 &0.036  & \cellcolor{red!15}100\%  &\cellcolor{red!15}0.029  & \cellcolor{red!15}100\%  &\cellcolor{red!15}0.062 & \cellcolor{red!15}100\%& \cellcolor{red!15}\red{0.020}&\cellcolor{red!15}100\%&\cellcolor{red!15}0.158&\cellcolor{red!15}100\%&\cellcolor{red!15}0.081&\cellcolor{red!15}100\%&\cellcolor{red!15}
${\blue{0.028}}_{(\blue{{\textbf{1.04}\times^\downarrow}})} $ \\ \hline
90&  10.5& 0.041 &\cellcolor{red!15}100\% & \cellcolor{red!15}\blue{0.111}&\cellcolor{blue!15}49.4\% & \cellcolor{blue!15}0.098* &\cellcolor{red!15}100\%  & \cellcolor{red!15}0.123 & \cellcolor{blue!15}47.8\%& \cellcolor{blue!15}0.481*  &\cellcolor{red!15}100\%  & \cellcolor{red!15}0.260 &\cellcolor{red!15}100\% & 
\cellcolor{red!15}$\red{0.084}_{(\blue{{\textbf{1.32}\times^\downarrow}})} $
\\ \hline
120& 13.8 & 0.042 & \cellcolor{red!15}100\%&\cellcolor{red!15}\blue{0.120}&\cellcolor{blue!15}36.7\% &\cellcolor{blue!15} $\infty$* &\cellcolor{red!15}100\% &\cellcolor{red!15}0.130& \cellcolor{blue!15}35.4\%&\cellcolor{blue!15}0.523*   &\cellcolor{red!15}100\%  & \cellcolor{red!15}0.272 & \cellcolor{red!15}100\% & \cellcolor{red!15}
$\red{0.098}_{(\blue{{\textbf{1.22}\times^\downarrow}})} $
\\ \hline
150& 15.6 & 0.044 & \cellcolor{red!15}100\%& \cellcolor{red!15}0.138& \cellcolor{blue!15}32.4\% & \cellcolor{blue!15}$\infty$* &\cellcolor{red!15}100\% & \cellcolor{red!15}\blue{0.134}&\cellcolor{blue!15}31.3\%&\cellcolor{blue!15}0.510*  &\cellcolor{red!15}100\%  & \cellcolor{red!15}{0.280} &\cellcolor{red!15}100\% & 
\cellcolor{red!15}$\red{0.112}_{(\blue{{\textbf{1.23}\times^\downarrow}})} $\\ \Xhline{1.2pt}
\end{tabular}
    \begin{tablenotes}
    \footnotesize
    \item[1. ] 1. \red{Red} text indicates the best result. \blue{Blue} text indicates the 2nd best result. Ratio = runtime / total dataset duration.
    \item[2. ] 2. \colorbox{red!15}{Red} background indicates SLAM ran successfully and completely. \colorbox{blue!15}{Blue} background indicates SLAM interrupted during the process.
    \item[3. ] 3. The ATE column represents the RMSE values. ${(\blue{{\textbf{value}\times^\downarrow}})}$ indicates the error reduction factor of Water-DSLAM compared to DP-INS subsystem.
    \end{tablenotes}
\end{table*}

We have previously validated the superiority of different subsystems underwater. However, due to the instability of underwater texture and structural features, a single subsystem struggles to maintain stable performance over large areas. To address this, we propose Water-DSLAM, a continuous and fault-tolerant multi-subsystem dense SLAM framework. To verify the contributions of each subsystem and the necessity of fault detection, we conducted systematic ablation experiments. In the experiments, "-N" denotes the absence, and "-F" the presence of the respective fault detection module.

As shown in Fig.~\ref{fig:b_4}(a), the Water-Scanner moves back and forth in the pool over a 150-second interval, with ground truth trajectories and structured light images recorded at 30~s, 90~s, 120~s and 150~s. 
Fig.~\ref{fig:b_4}(b) shows that DP-INS subsystem suffers from severe cumulative drift, while the performance of Water-Stereo and Water-UBSL subsystems varies depending on environmental features. In contrast, Water-DSLAM achieves more accurate and consistent trajectories through multi-subsystem fusion and fault-aware optimization. 
As shown in Table~\ref{tab:compare_pool}, by the final optimization step, Water-DSLAM, despite its initially slightly inferior performance, achieved the lowest ATE. This improvement, with an approximate 1.23-fold reduction in ATE, is attributed to the integration of constraints from multiple subsystems.
Without fault detection, Water-Stereo-N and Water-UBSL-N fail early due to poor texture or structural degradation with a continuity rate of around 35\%. However, the fault-tolerant Water-Stereo-F and Water-UBSL-F can detect failures and reinitialize using DP-INS, reaching a continuity rate of 100\%.

Fig. \ref{fig:b_4}(c)-(e) show the mapping results of the DP-INS, Water-UBSL, and Water-DSLAM, respectively. As seen, by utilizing structured light information, our Water-Scanner achieves dense 3D mapping underwater, a rare feat in such environments. Map from DP-INS exhibits ghosting and misaligned pool corners due to cumulative errors of acoustic odometry. 
While Water-UBSL leverages the IESKF framework to alleviate matching issues in structurally degraded areas, significant distortions still occur in regions lacking prominent features during large-scale motion, such as in the middle of a pool wall. 
In contrast, Water-DSLAM, by integrating different subsystems, ensures robust, uninterrupted underwater mapping with high consistency and minimal distortion.

\subsection{Comparison 1: Localized Fine-Scale Inspection}
\begin{figure*}[!htbp]
\centering
\includegraphics[width=1\linewidth]{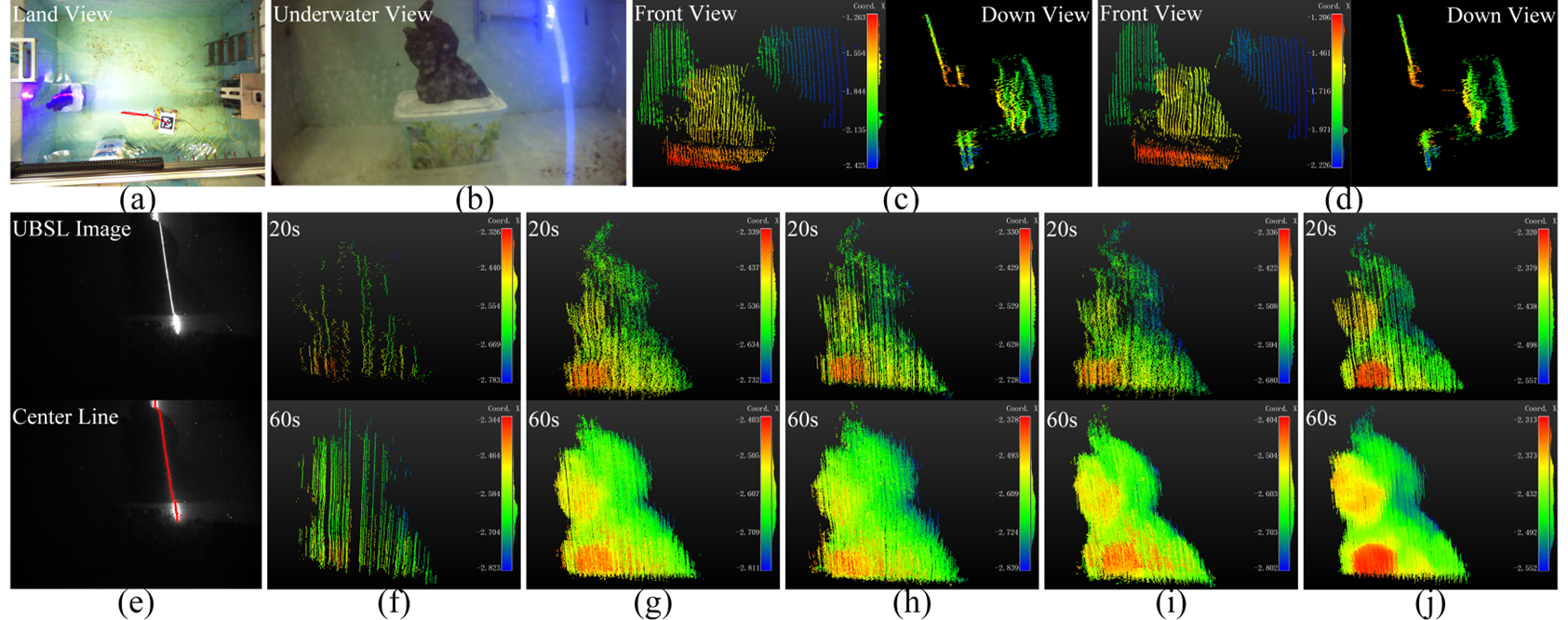}
\caption{{Comparative experimental results of state-of-the-art underwater dense mapping methods: self-stabilizing localized fine-Scale underwater inspection. (a)(b) Inspection scene. (c) Sweep pointcloud with no dekew \cite{palomer2019inspection}. (d) Sweep pointcloud with dekew in Water-DSLAM. (e) UBSL showcase. (f) Inspection results of \cite{bodenmann2017generation}. (g) Inspection results of \cite{ou2024structured}. (h) Inspection results of \cite{castillon2021extrinsic}. (i) Inspection results of DP-INS. (j) Inspection results of Water-DSLAM.}}
\label{fig:local_mapping}
\end{figure*}

High-precision localized fine-scale inspection is vital for many underwater tasks. However, vision-based methods often fail to provide dense data, while sonar-based approaches lack mapping accuracy. Built on the Water-DSLAM framework and our custom UBSL system, Water-Scanner enables accurate, close-range underwater observation. As shown in Fig.~\ref{fig:local_mapping}(a)(b), we demonstrate this capability through long-term observation of a underwater stone. The inspection effects of different existing SOTA methods are compared.

\subsubsection{Sweep Distortion}
To ensure accurate point cloud registration, we treat all line point clouds from a single scan as one sweep for improved matching consistency. Unlike previous methods that assume the robot remains stationary, our setup experiences inevitable vibrations from water-robot interactions, causing significant distortion in the sweep point cloud (Fig.~\ref{fig:local_mapping}(c)), which is often overlooked in prior work~\cite{palomer2019inspection}. Leveraging the high-frequency DP-INS, we obtain precise pose estimates for each line, effectively correcting distortion and producing clean, undistorted sweeps (Fig.~\ref{fig:local_mapping}(d)). This greatly reduces ghosting and underscores DP-INS’s advantage over low-frequency DR in structured light mapping.
\begin{figure*}[!htbp]
\centering
\includegraphics[width=0.9\linewidth]{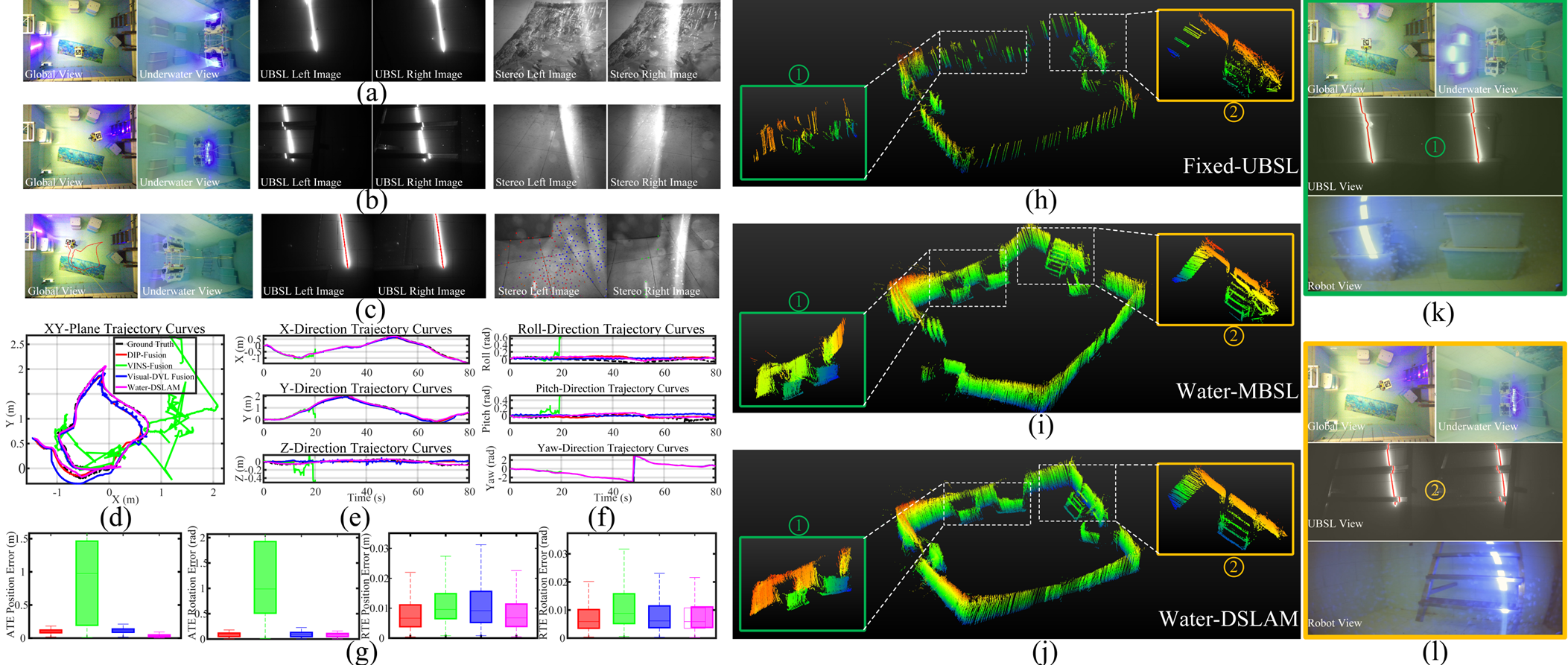}
\caption{{Comparative experimental results of state-of-the-art multi-modal underwater SLAM methods: random artificial pool dense mapping. (a) Underwater environment with strong textural features and weak structural features. (b) Underwater environment with weak textural features and strong structural features. (c) Demonstration of the randomized motion process of Water-Scanner. (d)-(f) Localization trajectory curves for different methods. (g) Localization errors of different methods. (h) Mapping results of Fixed-UBSL \cite{bodenmann2017generation,hitchcox2023improving}. (i) Mapping results of Water-MBSL \cite{ou2023water,palomer2019inspection}. (j) Mapping results of Water-DSLAM. (k)(l) correspond to the local regions marked as \textcircled{1}\textcircled{2}.}}
\label{fig:Pool_compare}
\end{figure*}
\begin{table*}[!htbp]
\label{tab:compare_pool}
{\caption{Quantitative Results of the Comparative Experiment on Different State-of-the-art Multi-modal Underwater SLAM Methods.\label{tab:compare_pool}}}
\centering
\begin{tabular}{c|@{\hspace{5pt}}c|@{\hspace{5pt}}c|@{\hspace{5pt}}c||c@{\hspace{5pt}}c|c@{\hspace{5pt}}c|c@{\hspace{5.5pt}}c@{\hspace{5.5pt}}|c@{\hspace{5.5pt}}c@{\hspace{5pt}}|c}
\Xhline{1.2pt}
\multirow{2}{*}{\begin{tabular}[c]{@{}c@{}}Dura.\\(s)\end{tabular}}& \multirow{2}{*}{\begin{tabular}[c]{@{}c@{}}Path\\(m)\end{tabular}}& \multirow{2}{*}{\begin{tabular}[c]{@{}c@{}}Ave. Vel.\\(m/s)\end{tabular}}&\multirow{2}{*}{Method}& \multicolumn{2}{c|}{ATE Pos. (m)}   & \multicolumn{2}{c|}{ATE Rot. (deg)}  & \multicolumn{2}{c|}{RTE Pos. (m)} & \multicolumn{2}{c|}{RTE Rot. (deg)} &\multirow{2}{*}{\begin{tabular}[c]{@{}c@{}}Continuity Ratio\end{tabular}}\\ 
     \cmidrule[0.5pt](rl){5-6}
    \cmidrule[0.5pt](rl){7-8}
    \cmidrule[0.5pt](rl){9-10}
    \cmidrule[0.5pt](rl){11-12}
 & & & &RMSE  & MAE  & RMSE   & MAE  &  RMSE    & MAE   &  RMSE  & MAE & \\ \Xhline{1.2pt}
\multirow{4}{*}{{80.8}}&\multirow{4}{*}{{13.0}}&\multirow{4}{*}{{0.096}} &\cellcolor{red!15}DIP-Fusion \cite{ou2024structured}&\cellcolor{red!15}\blue{0.068} &\cellcolor{red!15}\blue{0.063}&\cellcolor{red!15}7.762&\cellcolor{red!15}7.484&\cellcolor{red!15}\blue{0.019}  &\cellcolor{red!15}\blue{0.016} &\cellcolor{red!15}1.427& \cellcolor{red!15}1.101 &\cellcolor{red!15}\red{100\%}\\
 & & & \cellcolor{blue!15}VINS-Fusion\cite{qin2018vins,qin2019general}&\cellcolor{blue!15}{0.204}  &\cellcolor{blue!15}0.122  & \cellcolor{blue!15}16.242  &\cellcolor{blue!15}13.294   & \cellcolor{blue!15}0.099&\cellcolor{blue!15}0.042 &\cellcolor{blue!15}4.704   &\cellcolor{blue!15}2.241  &\cellcolor{blue!15}25.6\% (drift) \\
 & & & \cellcolor{blue!15}Visual-DVL Fusion\cite{huang2023tightly,xu2025aqua}& \cellcolor{blue!15}0.094  & \cellcolor{blue!15}0.083  & \cellcolor{blue!15}\blue{5.405}  &\cellcolor{blue!15}\red{4.731}   &\cellcolor{blue!15}0.031     &\cellcolor{blue!15}0.018 &\cellcolor{blue!15}\blue{1.348}   & \cellcolor{blue!15}\red{0.854}  &\cellcolor{blue!15}23.2\% (re-initialize) \\ 
 & & &\cellcolor{red!15} Water-DSLAM& \cellcolor{red!15}\red{0.039}  & \cellcolor{red!15}\red{0.032}  &\cellcolor{red!15}\red{5.330}  & \cellcolor{red!15}\blue{4.973}   &\cellcolor{red!15}\red{0.017}    & \cellcolor{red!15}\red{0.014} & \cellcolor{red!15}\red{1.309}   & \cellcolor{red!15}\blue{0.987} &\cellcolor{red!15}\red{100\%}\\ 
\Xhline{1.2pt}
\end{tabular}
    \begin{tablenotes}
    \footnotesize
    \item[1. ] 1. \red{Red} text indicates the best result. \blue{Blue} text indicates the 2nd best result.
    \item[2. ] 2. \colorbox{red!15}{Red} background indicates SLAM ran successfully and completely. \colorbox{blue!15}{Blue} background indicates SLAM interrupted during the process.
    \end{tablenotes}
\end{table*}
\subsubsection{Long-Time Scanning Mapping}
We further conducted a long-term in-situ fine-scale inspection of the underwater stone over a duration of 60 seconds, generating a dense 3D point cloud. 
To demonstrate the superiority of Water-DSLAM, we compared mapping results using various methods, including scan-based mapping with DR~\cite{bodenmann2017generation}, scan-based mapping with EKF~\cite{ou2024structured}, sweep-based mapping with distortion correction using DR~\cite{castillon2021extrinsic}, and sweep-based mapping using DP-INS. As shown in Fig.~\ref{fig:local_mapping}(e)–(j), the DR-based scan method loses significant data due to its low update rate, and although sweep-based DR offers slightly better density, it remains limited. Other methods can achieve higher density but suffer from severe ghosting caused by accumulated errors over time. Leveraging the constraints provided by the Water-UBSL and Water-Stereo subsystems, Water-DSLAM effectively mitigates the issue of cumulative errors, significantly enhancing mapping consistency. This result demonstrates the outstanding capability of Water-Scanner for high-precision, in-situ underwater inspection.

\subsection{Comparison 2: Free-Motion Dense Mapping}
To validate the superior performance of Water-DSLAM in uninterrupted localization and dense mapping, we conducted comparative experiments along large-scale free-motion trajectories within a real-world pool environment. To emulate the diverse feature distributions typically encountered during practical underwater operations, we arranged various artificial objects in the pool, creating regions with distinct visual and geometric characteristics, as shown in Fig.~\ref{fig:Pool_compare}(a)(b). The area in Fig.~\ref{fig:Pool_compare}(a) features strong textures beneficial for stereo vision, but lacks structural features due to a flat white wall. Conversely, the area in Fig.~\ref{fig:Pool_compare}(b) offers rich geometric structure for point cloud matching but contains sparse textures. This setup replicates typical underwater perception conditions, providing a controlled and challenging scenario for quantitative evaluation.

\begin{figure*}[!htbp]
\centering
\includegraphics[width=0.9\linewidth]{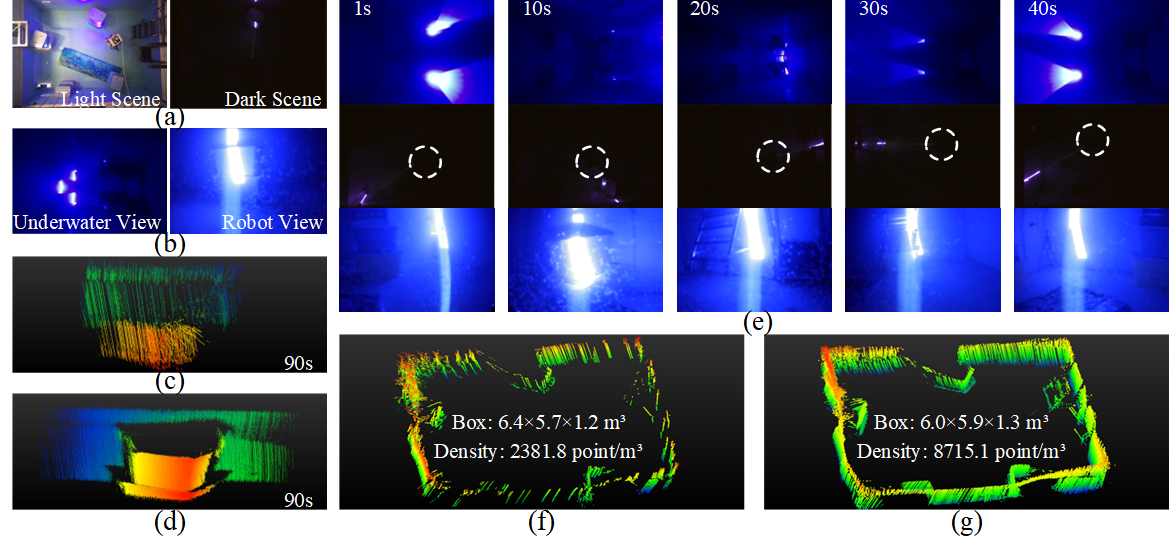}
\caption{Application experiments: underwater dark scene in-situ observation. (a)(b) Dark scene. (c) Self-stabilizing mapping as in \cite{palomer2019inspection}. (d) Self-stabilizing mapping by Water-DSLAM. (e) Snapshots of the image in motion mapping. (f) Motion mapping as in \cite{bodenmann2017generation,hitchcox2023improving}. (g) Motion mapping by Water-DSLAM.
}\label{fig:Dark_pool_mapping}
\end{figure*} 

\subsubsection{Trajectory Performance Evaluation}
To assess the localization performance of Water-DSLAM, we conducted comparisons against representative SOAT underwater multi-sensor fusion SLAM systems, including DIP-Fusion (DVL, IMU, and pressure sensor fusion)~\cite{ou2024structured}, VINS-Fusion (visual-inertial SLAM)~\cite{qin2018vins,qin2019general}, and Visual-DVL Fusion~\cite{huang2023tightly,xu2025aqua}. As illustrated in Fig.~\ref{fig:Pool_compare}(c), the robot followed a free-motion trajectory (red line), with corresponding structured light scans and stereo tracking results. The localization outputs of all methods are presented in Fig.~\ref{fig:Pool_compare}(d). Results show that VINS-Fusion, which relies heavily on texture-rich scenes, rapidly fails in low-texture areas, leading to drift and tracking loss. DIP-Fusion and Visual-DVL Fusion can maintain continuous operation via DVL-based compensation, but DIP-Fusion accumulates drift over time, while Visual-DVL Fusion—primarily visual and only secondarily acoustic—requires frequent reinitialization when visual features are sparse. In contrast, Water-DSLAM maintains seamless, drift-resistant localization with trajectories closely matching the ground truth, as further illustrated in Fig.~\ref{fig:Pool_compare}(e) and (f), which plot position and orientation estimates. These results highlight the robustness of Water-DSLAM’s tightly integrated multi-subsystem architecture.
Table~\ref{tab:compare_pool} reports the RMSE and MAE of ATE and RTE in both position and rotation. Water-DSLAM achieves the best performance across all metrics, with ATE position RMSE of 0.039~m and RTE position RMSE of 0.017~m, outperforming DIP-Fusion (0.068~m, 0.019~m) and Visual-DVL Fusion (0.094~m, 0.031~m). In rotation, Water-DSLAM yields an ATE RMSE of 5.30° and RTE RMSE of 1.31°, also lower than others. It maintains a 100\% continuity ratio, while VINS-Fusion and Visual-DVL Fusion show significant tracking loss with only 25.6\% and 23.2\% continuity, respectively. The box plots in Fig.~\ref{fig:Pool_compare}(g) further highlight Water-DSLAM’s lower median error and tighter distribution, indicating both high accuracy and robustness.

\subsubsection{Mapping Performance Evaluation}
Achieving large-scale dense reconstruction underwater remains a major challenge. While underwater structured light systems offer a promising solution, most existing methods rely on fixed-line configurations (Fixed-UBSL) with low-frequency DR, which limit data richness despite accurate 3D capture. In contrast, Water-Scanner, based on a scanning structured light system, enables denser reconstruction and introduces USBL constraints to enhance consistency. We compared Water-DSLAM with SOTA structured light approaches, including Fixed-UBSL~\cite{bodenmann2017generation, hitchcox2023improving} and Water-MBSL~\cite{ou2023water, palomer2019inspection}, as shown in Fig.~\ref{fig:Pool_compare}(h)–(j). Fixed-UBSL yields sparse results due to low data rates and poor odometry, making it unsuitable for detailed inspection. Water-MBSL improves density but exhibits severe distortion in regions with degraded structure, leading to reduced map consistency. In contrast, Water-DSLAM integrates stereo constraints for accurate localization, USBL for consistent mapping, and DP-INS for continuous motion tracking. \textcircled{1}\textcircled{2}) highlight local mapping results in two representative areas Fig.~\ref{fig:Pool_compare}(k)(l), where Water-DSLAM preserves fine structural details, demonstrating its effectiveness for large-scale, in-situ underwater observation.

\subsection{Application 1: Underwater Dark Scene In-situ Observation} 
In underwater environments, light absorption often leads to extremely low illumination, significantly limiting the applicability of passive vision-based SLAM methods. Water-Scanner addresses this challenge through an lased-aided structured light system, enabling effective 3D mapping in dark scenes. To validate its in-situ observation capability under such conditions, we conducted experiments in a real pool. Fig.~\ref{fig:Dark_pool_mapping}(a) respectively presents the underwater environment under normal lighting conditions and under near-total darkness at night. In this dark setting, stereo cameras fail to capture usable images, yet Water-DSLAM continues to operate robustly. Two experiments—self-stabilizing and motion observations—were conducted to comprehensively evaluate its performance under these low-light conditions.
The first is a self-stabilizing static observation experiment designed to evaluate Water-Scanner’s ability to capture fine structural details in dark underwater scenes. As shown in Fig.~\ref{fig:Dark_pool_mapping}(b), the robot operated in stabilization mode and scanned an underwater structure continuously for 90~s. The reconstructed results in Fig.~\ref{fig:Dark_pool_mapping}(c) show that, compared to~\cite{palomer2019inspection}'s method, which suffers from significant ghosting and blurring due to slight shaking over time, Water-DSLAM provides clearer and more accurate point clouds. Benefiting from the Water-UBSL subsystem, it preserves structural details effectively even in complete darkness, as shown in Fig.~\ref{fig:Dark_pool_mapping}(d). The second is a motion observation experiment that assesses the capability of large-scale in-situ mapping. Fig.~\ref{fig:Dark_pool_mapping}(e) presents visual perspectives captured at different locations, including underwater scenes and robot viewpoints. Although the visual subsystem was inactive under low-light conditions, the Water-UBSL subsystem maintained strong performance. As shown in Fig.~\ref{fig:Dark_pool_mapping}(f)(g), Water-DSLAM is compared with a SOTA baseline mapping method that integrates DR data with fixed-line structured light scans~\cite{bodenmann2017generation, hitchcox2023improving}. While both approaches benefit from the robustness of the active illumination system, only Water-DSLAM achieves dense mapping results comparable to those in well-lit environments. Its point cloud density reach 8715.1 point/$m^3$. Compared with the existing method, our system achieved a 3.7-fold improvement, demonstrating its strong adaptability and robustness in dark underwater scenarios.

\subsection{Application 2: Large-Scale Sinkhole In-situ Observation} 
\begin{figure*}[!htbp]
\centering
\includegraphics[width=0.9\linewidth]{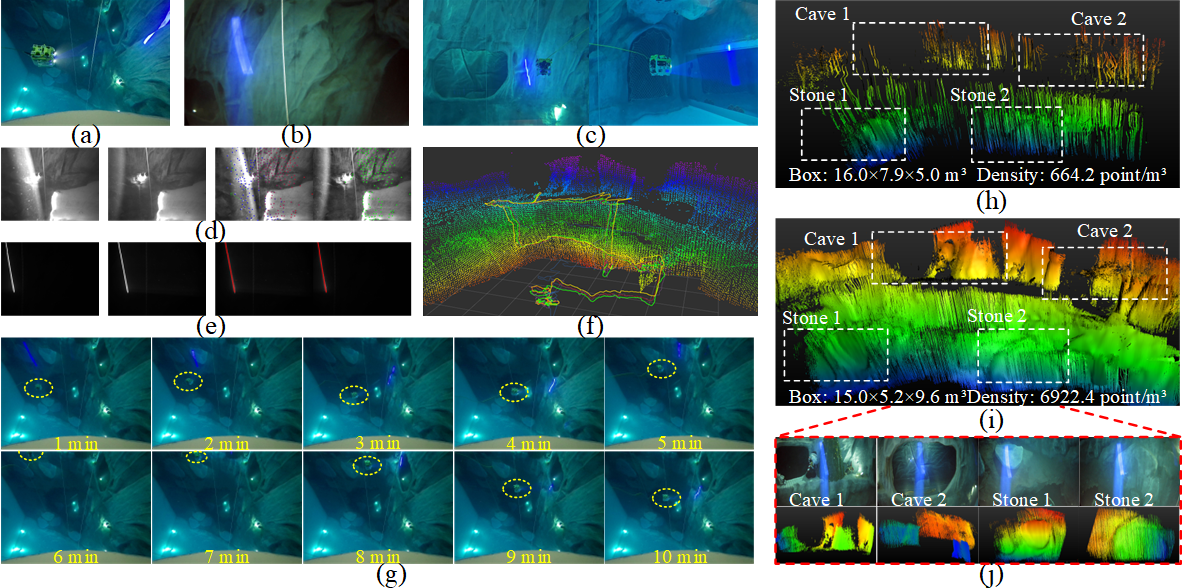}
\caption{Application experiments: underwater large-scale sinkhole in-situ
mapping. (a)-(c) 16-meter-deep sinkhole showcase. (d) Water-Stereo subsystem showcase. (e) Water-UBSL subsystem showcase.  (f) Large-scale in-situ observation trajectory of the Water-Scanner. (g) Snapshots of the image within 10 minutes. (h) Large-scale mapping results as in \cite{bodenmann2017generation,hitchcox2023improving}. (i) Large-scale mapping results by Water-DSLAM. (j) Local observation details of Water-DSLAM.
}
\label{fig:Qianshuiguan_mapping}
\end{figure*} 
To further validate the observation capabilities of Water-Scanner, we conducted a large-scale, long-duration dense mapping experiment in an underwater sinkhole at a depth of 16 meters.
The robot moves for up to 10 minutes and 750 $m^3$ space.
Compared to earlier pool experiments, this test involved longer operation time, wider coverage, and more realistic underwater conditions, thereby offering a more rigorous evaluation of Water-DSLAM.
As illustrated in Fig. \ref{fig:Qianshuiguan_mapping}(a)(b)(c), the sinkhole features complex structures including large stones and caves. The robot executed a 10-minute trajectory, ascending from the bottom while performing horizontal scanning at multiple depths using the Water-DSLAM framework. 
Fig. \ref{fig:Qianshuiguan_mapping}(d)(e) show the details of the Water-Stereo and Water-UBSL subsystems of the process, respectively.
The robot's motion trajectory is shown in Fig. \ref{fig:Qianshuiguan_mapping}(f).
In addition, Fig. \ref{fig:Qianshuiguan_mapping}(g) is a series of snapshots, which records the Water-Scanner's observed position every minute.
The final reconstruction results are presented in Fig. \ref{fig:Qianshuiguan_mapping}(h)(i), where Fig. \ref{fig:Qianshuiguan_mapping}(h) shows the result from the SOTA mapping method that integrates DR data with fixed-line structured light scans~\cite{bodenmann2017generation, hitchcox2023improving}, while Fig. \ref{fig:Qianshuiguan_mapping}(i) shows the output of our Water-DSLAM. Benefiting from high-frequency odometry and a robust algorithmic framework, our system achieved dense, high-fidelity mapping of the sinkhole, clearly capturing structural features such as caves and large boulders as shown in \ref{fig:Qianshuiguan_mapping}(j). 
Compared to the previous point cloud density of 664.2 point/$m^3$, Water-DSLAM achieved an observation result of 6922.4 points/$m^3$. The point cloud density increased by an order of magnitude, demonstrating the excellent in-situ observation capability of our system in long-duration and large-scale scenarios.

\subsection{Application 3: Field River In-situ Observation} 
\begin{figure*}[!htbp]
\centering
\includegraphics[width=0.9\linewidth]{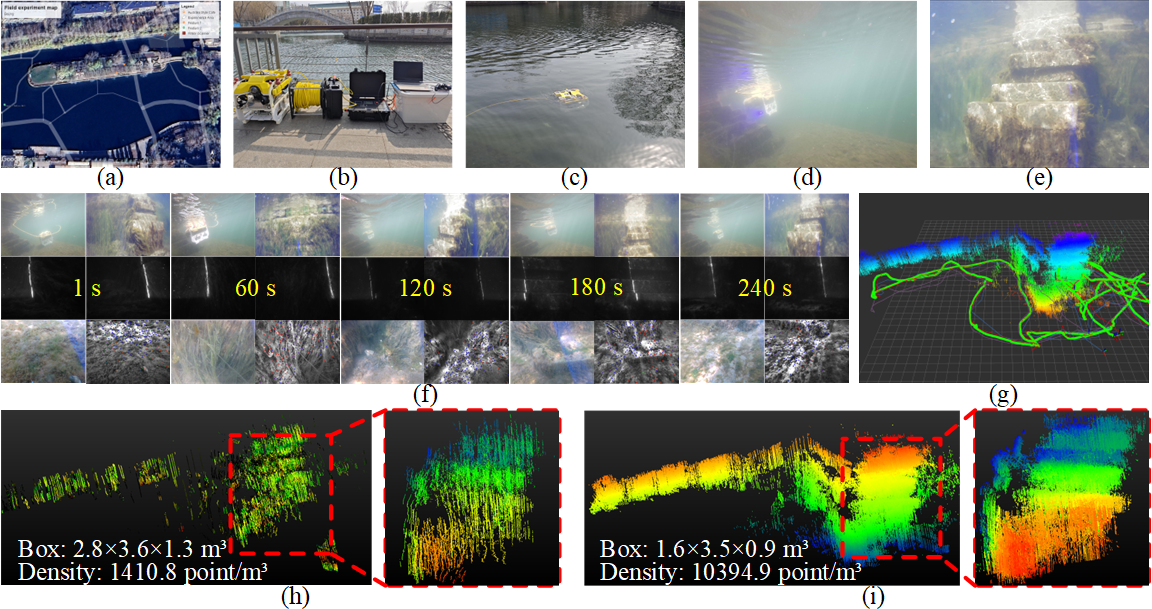}
\caption{Application experiments: Field river in-situ observation. (a) Observation site image derived from Google Earth. (b)-(d) Water-Scanner in field. (e) Primary in-situ field observation target.  (f) Snapshots of the image within 240 seconds. (g) Field in-situ observation trajectory of the Water-Scanner. (h) Field mapping results as in \cite{bodenmann2017generation,hitchcox2023improving}. (i) Field mapping results by Water-DSLAM. }
\label{fig:Field_mapping}
\end{figure*} 
To further assess the in-situ observation capabilities of Water-Scanner in more challenging settings, we conducted underwater tests at a field river, named Changhewan in Beijing, as illustrated in Fig. \ref{fig:Field_mapping}(a)-(e). 
In this field scenario, the robot's motion is significantly affected by water currents, and the effects of particles, light changes, water plants, and so on in the scene make the in-situ observation task more challenging.
As shown in Fig. \ref{fig:Field_mapping}(f), the robot performed autonomous localization and mapping along a riverbank wall and terrace structure over a 240-second trajectory depicted in Fig. \ref{fig:Field_mapping}(g).
Similarly, mapping results from the SOTA method that integrates DR with fixed-line structured light scans~\cite{bodenmann2017generation, hitchcox2023improving} (Fig. \ref{fig:Field_mapping}(h)) were compared with those from our Water-DSLAM system (Fig. \ref{fig:Field_mapping}(i)). As shown in the locally zoomed-in image, the existing method fails to reveal the details of the target structure, whereas our method clearly captures and preserves these structural features. 
It is worth noting that the point cloud density achieved by Water-DSLAM reached an impressive 10394.9 points/$m^3$.
The point cloud density increased by approximately 7-fold compared to the SOTA method, which achieved 1410.8 points/$m^3$.
These results validate Water-Scanner’s robustness in dynamic and unstructured aquatic environments and highlight its potential for dense in-situ mapping in field applications.

\section{Conclusion}
In this paper, we proposed an uninterrupted and fault-tolerant dense SLAM framework, named Water-DSLAM. To the best of our knowledge, this is the first structured-light-based underwater multi-modal dense SLAM system capable of continuous, fine-grained in-situ observation.
Water-DSLAM introduces a fault-tolerant triple-subsystem front-end architecture, which innovatively mitigates the adverse effects of random external sensor faults caused by environmental changes on the SLAM process. Additionally, a multi-modal factor graph-based back-end is integrated into Water-DSLAM to dynamically fuse heterogeneous sensor data. The proposed multi-sensor factor graph maintenance strategy and fault detection mechanisms efficiently manage asynchronous sensor frequencies and partial data loss.
To validate the effectiveness of Water-DSLAM, we developed Water-Scanner, a multi-sensor fusion robotics platform featuring a proprietary UBSL module that enables high-precision 3D perception. 
Experimental results demonstrate Water-DSLAM achieves superior robustness (0.039 m trajectory RMSE and 100\% continuity ratio during partial sensor dropout) and dense mapping (6922.4 points/$m^3$ in 750 $m^3$ water volume, approximately 10 times denser than existing methods.) across various challenging environments, including pools, dark underwater scenes, 16-meter-deep sinkholes, and field rivers.
Overall, Water-DSLAM represents a significant advancement in underwater robotics, offering a robust, accurate, and scalable solution for in-situ observation and dense mapping in real-world underwater scenarios.
It is expected to be applied to autonomous navigation tasks, including underwater exploration and coverage path planning.

\bibliographystyle{IEEEtran}

\bibliography{Water-DSLAM}

\end{document}